\documentclass{article}
\usepackage{iclr2027_conference,times}

\usepackage{amsmath}
\usepackage{amssymb}
\usepackage{booktabs}
\usepackage{graphicx}
\usepackage{placeins}
\usepackage{float}
\usepackage{hyperref}
\usepackage{url}

\iclrfinalcopy
\makeatletter
\def\@maketitle{\vbox{\hsize\textwidth
  {\LARGE\scshape\@title\par}
  \vskip 0.14in
  {\normalsize\@author\par}
  \vskip 0.22in}}
\makeatother
\renewcommand{\headrulewidth}{0pt}
\hypersetup{
  pdftitle={ControlTrace: Recovering Control Fields for Hidden-Content Recognition},
  pdfauthor={Zijian Liu, Yaoguang Chen, Liwei Liu, Weixi Wu, Hanming Zhang, Jiashui Wang, Na Ruan},
  pdfsubject={Hidden-content recognition and control-field recovery},
  pdfkeywords={Hidden-content recognition, Vision-language models, Conditional diffusion models, Control-field recovery, Image reconstruction},
  hidelinks
}

\title{ControlTrace: Recovering Control Fields\\
for Hidden-Content Recognition}

\author{%
Zijian Liu\textsuperscript{1}\quad
Yaoguang Chen\textsuperscript{2}\quad
Liwei Liu\textsuperscript{1}\quad
Weixi Wu\textsuperscript{2}\\[3pt]
Hanming Zhang\textsuperscript{2}\quad
Jiashui Wang\textsuperscript{2}\quad
Na Ruan\textsuperscript{1,*}\\[6pt]
{\small \textsuperscript{1}Shanghai Jiao Tong University\quad
\textsuperscript{2}Ant Group}\\[3pt]
{\small \textsuperscript{*}Corresponding author:
\href{mailto:naruan@sjtu.edu.cn}{\nolinkurl{naruan@sjtu.edu.cn}}}}

\newcommand{\method}{ControlTrace}
\newcommand{\net}{ControlTraceNet}
\newcommand{\best}[1]{\textbf{#1}}

\begin{document}

\maketitle
\fancyhead{}

\begin{abstract}
Spatially conditioned diffusion models can embed words and contours in
natural-looking images, but vision--language models (VLMs) may fail to
recognize the hidden content. Transformation-based recovery depends on
parameter and view selection. To evaluate hidden-content recovery and
recognition, we construct FreqBlind, a $6{,}000$-image benchmark spanning
contours, real words and non-words across three conditioning strengths.
The evaluated transformation-based methods show limited recognition of
contour patterns and weakly conditioned hidden content. To address this
limitation, we propose \method{} to recover the grayscale control field
used during generation. An $8.4$M-parameter U-Net predicts this field from
the carrier image, and a VLM then identifies its content. With
Qwen2.5-VL-7B-Instruct, \method{} achieves $60.2\%$ open-ended contour
recognition accuracy across the three conditioning strengths, exceeding
the best of the three evaluated prior methods by $26.9$ percentage points.
On an A100 GPU, the complete pipeline adds only $7.4$\,ms ($5.3\%$) to
direct VLM inference. Recovered fields have lower pixel errors and higher
structural similarity than the evaluated transformation views. Across four
evaluated VLMs, \method{} retains its overall contour recognition advantage.
Recognition remains stable under the tested JPEG compression, Gaussian
noise and downsampling. These results support control-field recovery for
hidden-content recognition in the evaluated setting.

\end{abstract}

\section{Introduction}
\label{sec:intro}

ControlNet-based diffusion pipelines can embed a rendered word, silhouette or
symbol within a natural-looking image \citep{zhang2023adding,qu2025hate}.
At low conditioning strength, the embedded structure may survive as faint
large-scale luminance variations within the generated image, which we call the
\textbf{carrier}. Squinting or viewing the carrier from a distance can reveal
this structure, yet a vision--language model (VLM), used here as a reader, may
fail to identify it.
Such images pose a challenge for content inspection, as both hidden text and
symbols can evade moderation models \citep{qu2025hate}.

Recent methods improve hidden-content recognition by transforming the carrier
before VLM recognition. SemVink \citep{li2025semvink} downsamples images,
SMSP \citep{tu2026smsp} combines low-pass views at multiple scales, and the
question-answering variant of Adaptive View Retrieval (AVR)
\citep{chen2026now} queries seven complementary views independently.
These strategies rely on choices of resolution, filtering and view
construction to suppress background detail while preserving hidden
structure.

\begin{figure}[!t]
\begin{center}
\includegraphics[width=\linewidth]{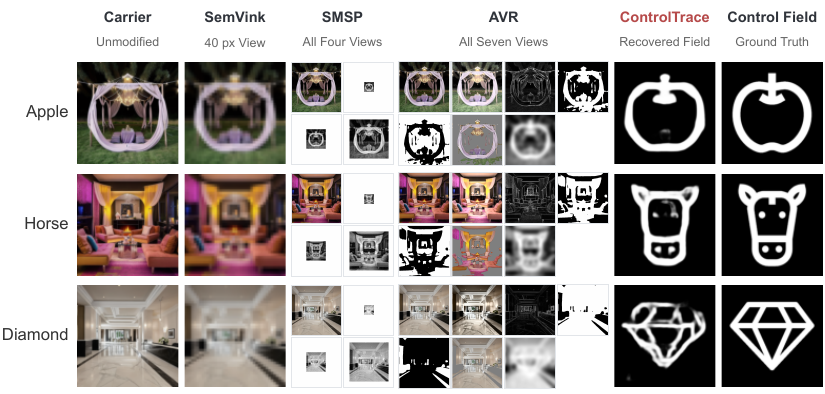}
\end{center}
\caption{Qualitative examples on the same carriers. SemVink shows its transformed
view; SMSP and AVR show all four and seven views, respectively. \method{}
predicts the control field, shown alongside the ground truth.}
\label{fig:teaser}
\end{figure}

To evaluate hidden-content recovery and recognition across target types
and conditioning strengths, we construct FreqBlind, a benchmark of contour
targets and rendered words and non-words. With
Qwen2.5-VL-7B-Instruct, our implementations of SemVink, SMSP and AVR
seven-view QA achieve $19.43\%$, $33.33\%$ and $19.37\%$ open-ended contour
recognition accuracy, respectively. AVR uses our fixed plurality readout;
its $45.10\%$ any-view coverage is a separate diagnostic that counts any
correct response among the seven views.

To address this limitation, we propose \method{}, which learns to recover
the grayscale \textbf{control field} supplied during image generation
(Figure~\ref{fig:framework}). This field directly specifies the hidden
structure and provides a spatial recovery target without the generated
scene texture (Figure~\ref{fig:teaser}). We train a U-Net \citep{ronneberger2015u} on paired carriers
and control fields, then pass the recovered field to a VLM for open-ended
recognition. This separates structural recovery from semantic interpretation
while requiring only the carrier image for recovery at inference.

Across the three conditioning strengths, \method{} achieves $60.2\%$
open-ended contour accuracy on FreqBlind, exceeding SMSP, the best of the
three evaluated prior methods, by $26.9$ percentage points and AVR any-view
coverage by $15.1$ points. The gain over SMSP is largest under weak
conditioning. Direct comparisons with the generating control fields show
lower pixel errors and higher structural similarity than the evaluated
transformed views in both domains. With Qwen2.5-VL-7B-Instruct,
\method{} requires a single VLM call with a small measured latency
increase over direct inference.

Our contributions are as follows:

\begin{itemize}
\setlength{\itemsep}{1pt}
\item We construct FreqBlind, a test dataset for hidden-content recovery containing
      contour targets and rendered words and non-words across three
      conditioning strengths.
\item We propose \method{}, which learns to recover the generator's grayscale
      control field from a carrier image. A VLM then recognizes hidden content
      in the recovered field.
\item We evaluate reconstruction quality, recognition, inference efficiency and
      component ablations, together with generalization across VLM readers
      and generators and robustness to image perturbations. These experiments
      show improved contour recognition, especially under weak conditioning,
      and identify limitations in word recognition and cross-generator transfer.
\end{itemize}

\section{Related Work}
\label{sec:related}

\paragraph{Hidden-content generation and evaluation.}
Diffusion models with spatial conditioning can embed text and shapes within
natural-looking scenes. \citet{qu2025hate} used Stable Diffusion and
ControlNet to construct Hateful Illusion and demonstrated failures of moderation
models and VLMs on hidden hateful content. HC-Bench \citep{li2025semvink}
evaluates recognition of hidden text and objects, whereas IlluChar
\citep{tu2026smsp} examines hidden-character perception across pattern scales.
These benchmarks establish hidden-content recognition as a distinct challenge
for visual understanding.

\paragraph{Hidden-content recognition.}
SemVink \citep{li2025semvink} downsamples carriers to emphasize global
structure, whereas SMSP \citep{tu2026smsp} combines low-pass filtering and
spatial rescaling to expose patterns at different scales. AVR
\citep{chen2026now} learns to weight seven complementary views for
hidden-message template retrieval. Its open-ended VLM protocol questions
these views independently; we evaluate this QA variant, without the
template bank or learned retrieval gate. These approaches expose hidden
structure through carrier transformations and view construction. Our
comparison asks whether learning to recover the underlying control field
provides a more effective reconstruction objective.

\paragraph{Learned recovery of visual encodings.}
NeuralMagicEye \citep{zou2020neuralmagiceye} learns to recover depth encoded by
texture disparities in autostereograms. Its watermarking experiments encode
character and QR-code patterns as depth, blend the resulting autostereograms
with background carrier images, and recover the patterns. This is related to
our use of paired synthetic data to train a recovery network. The encoding
mechanisms and reconstruction targets differ: NeuralMagicEye decodes depth
represented by texture disparities, whereas \method{} predicts the grayscale
control field from a ControlNet-generated carrier. A VLM then performs
open-ended recognition on the recovered field.

\section{Problem Setting}
\label{sec:setting}

\paragraph{Task.}
A carrier image $I$ is generated from a scene prompt and a grayscale control
field $c$ by a text-to-image diffusion model with ControlNet. The contours or
text encoded in $c$ are hidden within the generated scene. Conditioning
strength $s$ controls the field's influence on generation. Given only $I$,
the task is to identify the hidden content represented by $c$ through an
open-ended response, without a candidate list or target template.

\paragraph{Training and inference conditions.}
For learned recovery, we retain the control fields used to generate the
training carriers. Training and FreqBlind share the SD1.5 architecture and
QR Code Monster encoder, but use different base checkpoints and sampling
budgets (Appendix~\ref{app:generation}). At test time, the control field
and generation prompt are unavailable. Inference requires only the carrier
image and a recognition prompt, with no generator call.

\paragraph{FreqBlind benchmark.}
We construct FreqBlind with $50$ contour targets and $50$ word targets.
Each target is rendered with $20$ scene prompts drawn from a pool of $200$, at three
conditioning strengths $s \in \{1.0, 1.5, 2.0\}$, yielding $6{,}000$ carriers.
The contour targets are named by English common nouns, excluding proper nouns
and landmarks. The word targets comprise $25$ real four-letter words and $25$
pronounceable non-words of the same length. Word-domain scores therefore
combine equal lexical and non-lexical halves.

Carriers use Realistic Vision V5.1, a Stable Diffusion~1.5 checkpoint
\citep{rombach2022high}, with QR Code Monster ControlNet. Both domains share
ordinary indoor and outdoor scene prompts. Conditioning strength $s$
scales the control branch's contribution during generation.

\section{\method{}}
\label{sec:method}

\subsection{Control-Field Reconstruction}
\label{sec:reconstruction}

\begin{figure}[!t]
\centering
\includegraphics[width=\linewidth]{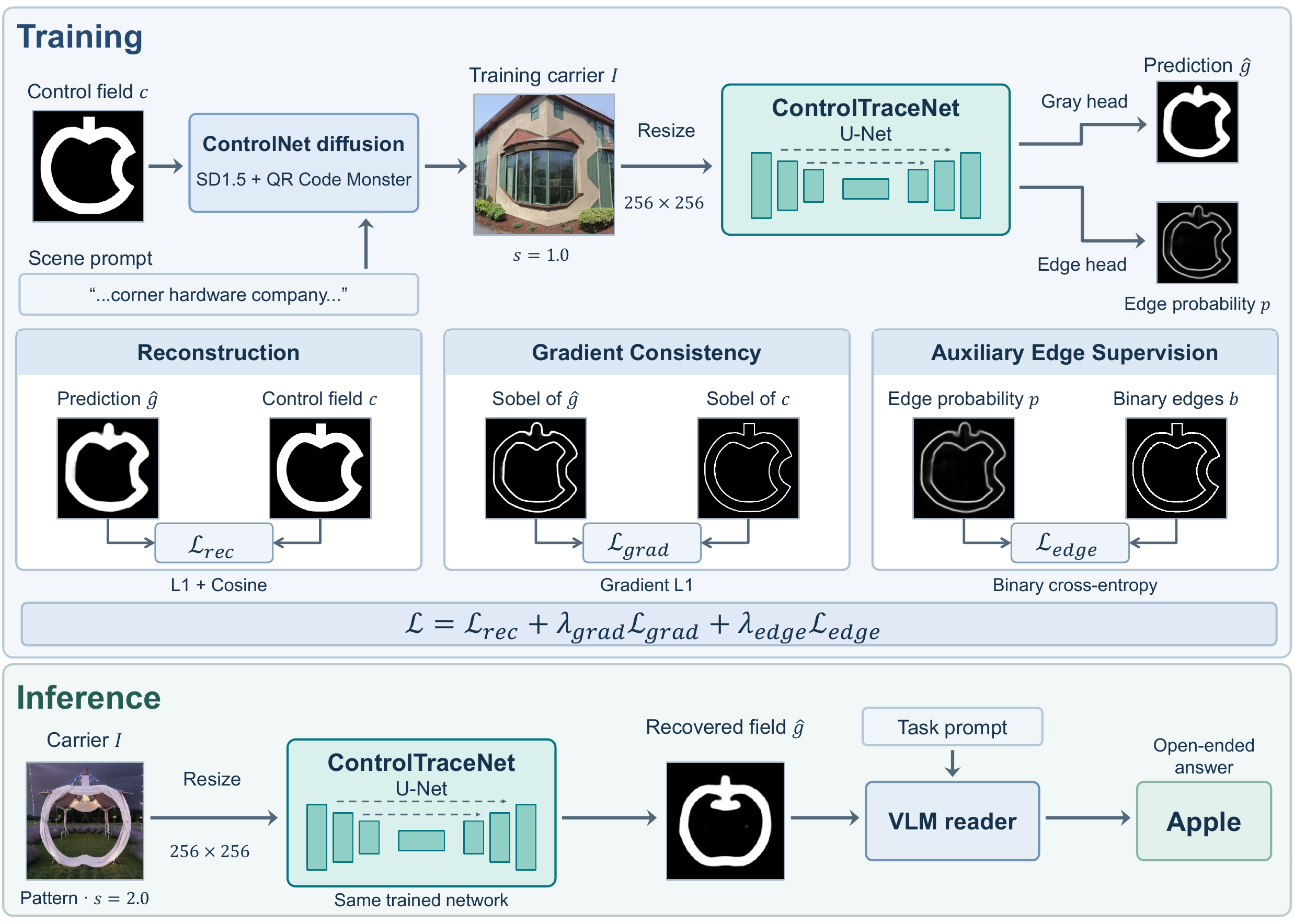}
\caption{ControlTrace workflow and training losses. Reconstruction and gradient losses supervise the recovered control field; the auxiliary edge loss is used only during training. At inference, a VLM reads the recovered field with a task prompt.}
\label{fig:framework}
\end{figure}

Transforming a carrier requires balancing scene-texture suppression against
retention of hidden structure. We therefore predict the generating control
field, which directly specifies the target's spatial structure. For a given
control field, different scene prompts produce different carrier appearances
while retaining the same spatial target. Retained carrier--control pairs
therefore provide a common supervision signal across these variants, without
requiring the recovery network to reproduce scene-dependent textures
(Figure~\ref{fig:framework}).

To combine spatial context with boundary detail, \net{} uses a four-level
U-Net with skip connections between corresponding encoder and decoder scales
\citep{ronneberger2015u}. The $8.37$M-parameter network predicts a grayscale
field $\hat{g}$ and auxiliary edge logits $\hat{e}$ from shared decoder features
(Appendix~\ref{app:training}).

At inference, a $512\times512$ carrier is resized to $256\times256$ for recovery.
A VLM reads the grayscale field $\hat{g}$ with the task prompt in
Appendix~\ref{app:prompts} to produce an open-ended answer; auxiliary edge
predictions are discarded.

\subsection{Training Objective}
\label{sec:objective}

\paragraph{Reconstruction.}
For target $c\in[0,1]^N$ and prediction $\hat{g}\in[0,1]^N$ with $N$ pixels,
L1 penalizes absolute grayscale error, while cosine measures normalized
alignment across pixels. Fields are flattened, with stabilized norms
$d(x)=\max\{\|x\|_2,\varepsilon_c\}$:
\begin{equation*}
\mathcal{L}_{\mathrm{rec}}
 = \frac{1}{N}\sum\nolimits_{i=1}^{N}|\hat{g}_i-c_i|
    +1-\frac{\hat{g}^{\mathsf T}c}{d(\hat{g})d(c)},
\end{equation*}

The L1 term penalizes absolute intensity differences that cosine alignment
alone cannot resolve. The cosine term compares normalized grayscale values at
corresponding pixels, adding a field-level alignment signal to the pixelwise
penalty.

\paragraph{Gradient consistency.}
Reconstruction does not explicitly constrain local boundary transitions in the
field. We therefore compare Sobel gradient magnitudes $E(x)$, clipped to $[0,1]$,
directly on the recovered and target fields:
\begin{equation*}
\mathcal{L}_{\mathrm{grad}}
 = \frac{1}{N}\sum\nolimits_{i=1}^{N}|E(\hat{g})_i-E(c)_i|.
\end{equation*}
\paragraph{Auxiliary edge supervision.}
The auxiliary head adds edge-location supervision to the shared decoder
features. It predicts $p_i=\operatorname{sigmoid}(\hat{e}_i)$ for binary labels
$b_i=\mathbf{1}[E(c)_i>\tau]$, using
\begin{equation*}
\mathcal{L}_{\mathrm{edge}}
 = -\frac{1}{N}\sum\nolimits_{i=1}^{N}
    \big[b_i\log p_i+(1-b_i)\log(1-p_i)\big].
\end{equation*}
Gradient magnitudes do not determine the grayscale field itself, so gradient
consistency complements rather than replaces reconstruction. The auxiliary
head separately predicts edge locations from shared features; its supervision
acts on the decoder representation, while the gradient loss acts directly on
the grayscale output. Their weighted combination gives the per-example
objective:
\begin{equation}
\mathcal{L}
 = \mathcal{L}_{\mathrm{rec}}
   +\lambda_{\mathrm{grad}}\mathcal{L}_{\mathrm{grad}}
   +\lambda_{\mathrm{edge}}\mathcal{L}_{\mathrm{edge}}.
\label{eq:training_loss}
\end{equation}
Losses are averaged over the batch; Appendix~\ref{app:training} specifies the
Sobel operator and numerical stabilization, including cross-entropy from logits.

\section{Experiments}
\label{sec:experiments}
\renewcommand{\bottomfraction}{0.5}
\setcounter{bottomnumber}{1}

\subsection{Experimental Setup}
\label{sec:setup}

\paragraph{Evaluation protocol.}
FreqBlind recognition used Qwen2.5-VL-7B-Instruct \citep{bai2025qwen25vltechnicalreport} with $1{,}000$ carriers per domain--strength cell. Our implementations use SemVink's $40$-pixel view \citep{li2025semvink}, SMSP's three low-pass views plus the original \citep{tu2026smsp}, and AVR's seven independently queried views with fixed plurality voting \citep{chen2026now}. Separate from single-answer accuracy, \textbf{any-view coverage} measures the reference-scored fraction with at least one correct response. Appendix~\ref{app:avr} specifies AVR parameters and voting.

\paragraph{Training data and optimization.}
The data pool is built from $2{,}000$ FIGR-8 control images
\citep{clouatre2019figr} and $750$ scene prompts. The reference model uses
the $12{,}000$ carrier--control pairs generated at $s=1.0$.
Splitting by source control gives $9{,}600$ training, $1{,}200$ validation
and $1{,}200$ internal test pairs, shared across three seeds.
We train from scratch for $30$ epochs at learning rate $2\times10^{-4}$,
with $\lambda_{\mathrm{grad}}=0.25$, $\lambda_{\mathrm{edge}}=0.5$
and edge threshold $\tau=0.12$, selecting checkpoints by validation loss.
FreqBlind evaluates all three strengths with controls and scene prompts
disjoint from training (Appendix~\ref{app:factorial}).

\paragraph{Perturbation protocol.}
We evaluate \method{} under image perturbations using $300$ carriers per
domain--strength cell and one checkpoint. Perturbations are applied to
the carrier before recovery. A white-box $20$-step PGD attack uses the
recovery network and true control field to maximize control-field
reconstruction error within an $\ell_\infty$ budget on the recovery input
grid, without optimizing through the VLM.

\paragraph{Reconstruction metrics.}
We compare outputs with their paired control fields using mean squared
error (MSE), mean absolute error (MAE), and structural similarity (SSIM)
\citep{wang2004image}. Evaluation uses grayscale fields on a common
$256\times256$ grid in $[0,1]$. We average metrics across each method's
views or training seeds before averaging images; Appendix~\ref{app:quality}
specifies geometry and aggregation.

\paragraph{Efficiency protocol.}
We measured latency and token use on $300$ FreqBlind carriers spanning all target types and strengths, with two repetitions per method on one A100 80GB GPU. All methods used Qwen2.5-VL-7B-Instruct, batch size one, and greedy decoding capped at $128$ tokens; \method{} used one checkpoint. After warmup, synchronized timing covered preprocessing through the final answer, excluding model loading and disk I/O. SMSP used one four-image request; AVR used seven serial queries and voting.

\paragraph{Metrics and uncertainty.}
All methods use lowercase ASCII normalization and target-substring matching or at least $80\%$ token coverage for multi-token targets. Trained methods' main recognition and fidelity results average three seeds. Main AVR comparisons bootstrap target items $20{,}000$ times, retaining scenes and using trained-model scores averaged per record \citep{efron2000introduction}. These $95\%$ intervals condition on selected models and configurations and differ from seed standard deviations. Appendices~\ref{app:prompts}, \ref{app:judge} and~\ref{app:hcbench} provide prompts, alternative scoring and HC-Bench protocols.

\subsection{Recognition Performance}
\label{sec:main}

Across strengths, \method{} achieved $60.21\%$ contour accuracy, versus $19.43\%$ for SemVink, $33.33\%$ for SMSP and $19.37\%$ for AVR plurality (Table~\ref{tab:main}). The gain over SMSP, the strongest of these baselines, was $26.88$ points. Returning one answer, \method{} also exceeded AVR any-view coverage ($45.10\%$) by $15.11$ points (paired item-bootstrap $95\%$ CI, $[9.17,21.21]$).

\begin{table}[!t]
\centering
\small
\caption{Open-ended recognition on FreqBlind (\%), with $1{,}000$ carriers per domain--strength cell. Trained methods average three seeds; $^\dagger$ denotes our implementations. AVR QA uses fixed plurality voting; AVR any-view coverage is a reference-scored seven-answer diagnostic. Bold numbers mark the highest single-answer accuracy.}
\label{tab:main}
\begin{tabular}{lrrrr@{\hskip 1em}rrrr}
\toprule
& \multicolumn{4}{c}{Contour $\uparrow$} & \multicolumn{4}{c}{Word $\uparrow$} \\
\cmidrule(lr){2-5}\cmidrule(lr){6-9}
Method & All & $1.0$ & $1.5$ & $2.0$ & All & $1.0$ & $1.5$ & $2.0$ \\
\midrule
Carrier, read directly & 6.30 & 1.1 & 5.5 & 12.3 & 2.57 & 0.1 & 0.7 & 6.9 \\
SemVink$^\dagger$ (training-free) & 19.43 & 2.8 & 18.1 & 37.4 & 53.60 & 9.1 & 68.8 & 82.9 \\
SMSP$^\dagger$ (training-free) & 33.33 & 12.7 & 40.1 & 47.2 & \best{63.37} & 17.6 & \best{80.2} & \best{92.3} \\
AVR seven-view QA$^\dagger$ & 19.37 & 2.9 & 18.5 & 36.7 & 23.03 & 0.6 & 17.1 & 51.4 \\
AVR any-view coverage & 45.10 & 19.2 & 49.6 & 66.5 & 43.00 & 5.8 & 44.1 & 79.1 \\
NAFNet (trained, matched) & 56.51 & 38.1 & 62.8 & 68.6 & 56.76 & 19.3 & 70.8 & 80.2 \\
\midrule
\textbf{\method{} (Ours)} & \best{60.21} & \best{45.4} & \best{65.6} & \best{69.6} & 61.62 & \best{28.3} & 75.1 & 81.5 \\
\bottomrule
\end{tabular}
\end{table}

The gain over SMSP peaked at $s=1.0$, where \method{} achieved $45.4\%$, versus $12.7\%$ for SMSP and $19.2\%$ AVR coverage. The gap to AVR coverage narrowed to $3.1$ points at $s=2.0$, with a $95\%$ interval containing zero. Figure~\ref{fig:teaser} contrasts scene-retaining transformed views with recovered outlines.

Sweeping Gaussian blur and downsampling isolated parameter sensitivity from complete pipeline comparisons. Both operations peaked inside the tested grids, consistent with a balance between suppressing scene texture and preserving hidden structure. Their best tested settings still trailed \method{} on weak contours (Appendix~\ref{app:analysis}).

Word recognition followed a different pattern. The aggregate scores of
\method{} and SMSP were $61.62\%$ and $63.37\%$, with an interval for
their difference containing zero. Although \method{} led at the weakest
strength, SMSP reached $92.3\%$ at the strongest, compared with $81.5\%$
for \method{}. The principal advantage therefore concerns contour
recovery, rather than every target type or conditioning strength.

\paragraph{Lexical effects.}
\label{sec:lexical}

To examine lexical effects, we generated $1{,}500$ carriers from exact
anagrams of real words, preserving each pair's letters, font and rendering
procedure.
Replacing words with anagrams reduced aggregate single-answer accuracy by
$34.8$ points for SemVink, $23.2$ for SMSP, $24.3$ for AVR plurality and
$31.8$ for \method{}. AVR any-view coverage decreased by $30.5$ points. This supports a lexical contribution to recognition, although
the changed spatial arrangement prevents attributing the entire difference
to lexical completion. Appendix~\ref{app:lexicalanalysis} presents the paired
results and separates them from the benchmark's unmatched non-words.

\subsection{Recovery Quality}
\label{sec:nafnet}

We compared method outputs directly with their generating control fields
to measure recovery before VLM recognition (Table~\ref{tab:quality_main}).
On contours, \method{} achieved an MSE of $0.0327$ and SSIM of $0.8312$,
compared with $0.1172$ and $0.1193$ for SMSP's view mean.
The paired MSE reduction was $0.0846$ ($95\%$ item-bootstrap CI,
$[0.0726,0.0929]$).

\begin{table}[!t]
\centering
\small
\setlength{\tabcolsep}{5pt}
\caption{Control-field fidelity on FreqBlind ($3{,}000$ carriers per domain). Trained methods average three seeds; SMSP and AVR average four and seven view scores per carrier after fixed geometric mapping. Bold marks the best point estimate in each column.}
\label{tab:quality_main}
\begin{tabular}{lrrrrrr}
\toprule
& \multicolumn{3}{c}{Contour} & \multicolumn{3}{c}{Word} \\
\cmidrule(lr){2-4}\cmidrule(lr){5-7}
Method & MSE $\downarrow$ & MAE $\downarrow$ & SSIM $\uparrow$ & MSE $\downarrow$ & MAE $\downarrow$ & SSIM $\uparrow$ \\
\midrule
Carrier & 0.1049 & 0.2732 & 0.1259 & 0.1210 & 0.2966 & 0.0621 \\
SemVink & 0.1038 & 0.2859 & 0.1276 & 0.1168 & 0.3042 & 0.0631 \\
SMSP (view mean) & 0.1172 & 0.2960 & 0.1193 & 0.1401 & 0.3247 & 0.0583 \\
AVR (view mean) & 0.2579 & 0.3832 & 0.1710 & 0.2678 & 0.3935 & 0.1346 \\
Gaussian blur ($\sigma=16$) & 0.1252 & 0.3234 & 0.1096 & 0.1278 & 0.3254 & 0.0479 \\
NAFNet & 0.0341 & 0.0508 & 0.8259 & 0.0365 & 0.0559 & 0.8039 \\
\midrule
\textbf{\method{} (Ours)} & \best{0.0327} & \best{0.0496} & \best{0.8312} & \best{0.0319} & \best{0.0493} & \best{0.8272} \\
\bottomrule
\end{tabular}
\end{table}

\method{} had lower MSE and MAE and higher SSIM than the evaluated transformation summaries in both domains. However, SMSP retained higher aggregate word recognition (Table~\ref{tab:main}), separating field fidelity from semantic recognition. Multi-view fidelity averages characterize the available views rather than the selected answer (Appendix~\ref{app:quality}).

With the same U-Net, data and training budget, control-field supervision
achieved $45.40\%$ weak-contour accuracy, compared with $13.27\%$ for
stronger-conditioning scene supervision. The losses were shared without
target-specific retuning (Appendix~\ref{app:targets}). Replacing the U-Net
with NAFNet \citep{chen2022simple} at matched approximate capacity yielded
$56.51\%$ aggregate contour recognition, $3.70$ points below the U-Net
($95\%$ CI, $[2.39,5.10]$). Both backbones exceeded SMSP.

\paragraph{Component ablations.}
\label{sec:ablation}

Training allocation had a larger effect than additional architectural
complexity (Table~\ref{tab:ablation_main}). Concentrating training on $s=1.0$ improved weak-contour
accuracy by $6.60$ points relative to uniform allocation across strengths.
Removing the gradient loss reduced accuracy by $2.20$ points, whereas
removing the auxiliary edge head changed it by $0.27$ points.
Greater depth provided no consistent benefit. Appendix~\ref{app:factorial} gives the training-data
and input settings.
Configuration selection used downstream recognition; within each run,
checkpoint selection used validation loss (Appendix~\ref{app:pixel}).

\begin{table}[!t]
\centering
\renewcommand{\arraystretch}{1.2}
\caption{Component ablations on weak contours ($s=1.0$), averaged over three training seeds. Values are changes in recognition accuracy, in percentage points, relative to the corresponding reference configuration. Training-data and input settings are given in Appendix~\ref{app:factorial}.}
\label{tab:ablation_main}
\begin{tabular}{lr}
\toprule
Change & $\Delta$ accuracy (pp) \\
\midrule
Uniform training allocation across strengths & $-6.60$ \\
Remove gradient loss & $-2.20$ \\
Remove auxiliary edge head & $-0.27$ \\
Increase U-Net depth to five levels & $-1.13$ \\
Increase U-Net depth to six levels & $-1.07$ \\
\bottomrule
\end{tabular}
\end{table}

\subsection{Generalization and Robustness}
\label{sec:readers}

\begin{figure}[!t]
\centering
\includegraphics[width=\linewidth]{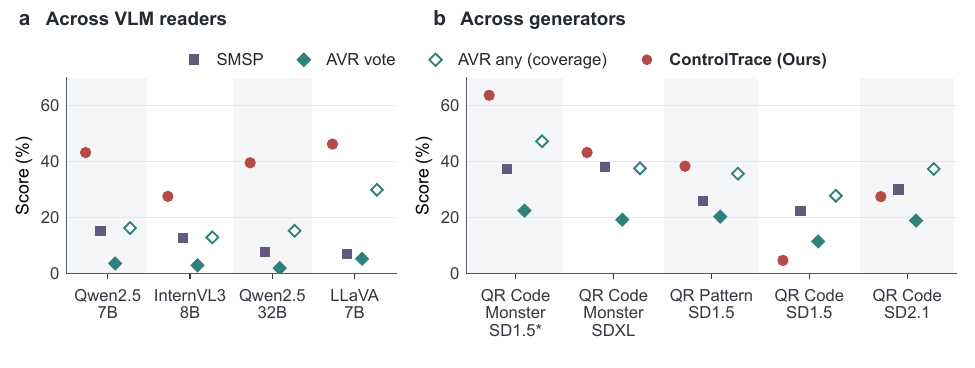}
\caption{Contour recognition across readers and generators. (a) Four readers on $300$ weak-contour carriers each. (b) Five QR-based generator configurations, each with $900$ carriers across strengths; the asterisk marks the training architecture and control encoder. ControlTrace uses one checkpoint. AVR any denotes reference-scored coverage across seven answers.}
\label{fig:generalization_main}
\end{figure}

\paragraph{VLM readers.}
We held the recovery model fixed and evaluated four readers on the same
$1{,}800$-carrier subset: Qwen2.5-VL-7B, InternVL3-8B~\citep{zhu2025internvl3}, Qwen2.5-VL-32B,
and LLaVA-OneVision-7B~\citep{li2024llava}. At weak conditioning, \method{} exceeded both
SMSP and AVR plurality on contours for every reader. Its gains over SMSP
were $15.0$--$39.3$ points, with each paired interval excluding zero
(Figure~\ref{fig:generalization_main}a).
Its gains over AVR any-view coverage ranged from $14.7$ to $27.0$ points.
Aggregate contour scores favored \method{} for every reader, although
individual cells reversed this ordering: InternVL3-8B at $s=2.0$ reached
$49.0\%$, below AVR coverage of $56.3\%$. Word results were more
reader-dependent (Appendix~\ref{app:readers}).

\paragraph{Generating models.}
Without adaptation, \method{} exceeded SMSP in aggregate contour accuracy on three of five QR-based configurations: QR Code Monster with SD1.5 and SDXL, and QR Pattern with SD1.5 (Figure~\ref{fig:generalization_main}b). Both QR Code configurations reversed this ordering. On SD1.5 with QR Code, accuracy was $4.78\%$, versus $22.56\%$ for SMSP and $27.89\%$ AVR coverage. Transfer therefore depended on the control encoder (Appendix~\ref{app:transfer}).

\begin{figure}[!t]
\centering
\includegraphics[width=\linewidth]{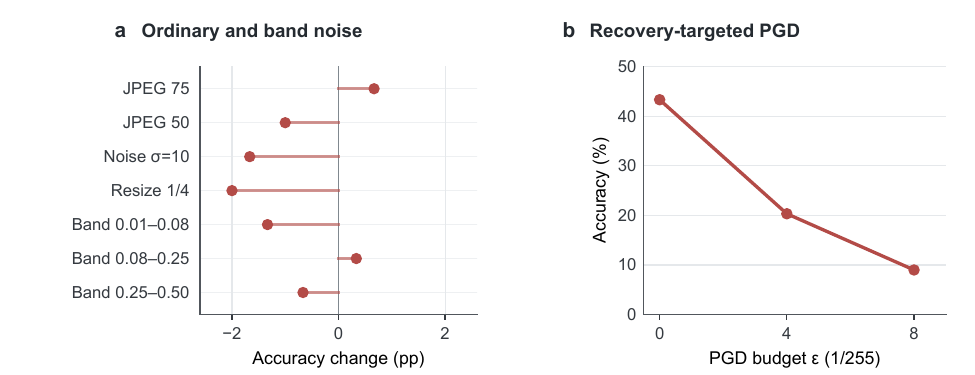}
\caption{ControlTrace under image perturbations, using $300$ weak-contour carriers and one checkpoint. (a) Accuracy changes under redistribution and band-limited noise. (b) Accuracy under PGD targeting the recovery network.}
\label{fig:robustness_unified}
\end{figure}

\paragraph{Image perturbations.}
JPEG compression, Gaussian noise and downsampling changed weak-contour
accuracy by at most $2.0$ points; band-limited noise changed it by at most
$1.33$ points (Figure~\ref{fig:robustness_unified}a).
Across strengths, the three redistribution operations changed average
accuracy by at most one point in both domains (Appendix~\ref{app:robust}).

By contrast, $20$-step PGD \citep{madry2017towards} reduced weak-contour
accuracy from $43.33\%$ to $20.33\%$ at $\varepsilon=4/255$ and
$9.00\%$ at $8/255$ (Figure~\ref{fig:robustness_unified}b).
The attack impaired recovery at every strength, although post-attack contour
recognition remained higher under stronger conditioning.
Because PGD targets reconstruction without differentiating through the VLM,
this decline identifies a vulnerability in the recovery stage
(Appendix~\ref{app:adversarial}).

\subsection{Specificity}
\label{sec:specificity}

Using a shared none-capable prompt, we evaluated hidden contours and four
negative sets: COCO photographs~\citep{lin2014microsoft}, matched control-off generations, dense-texture
scenes and non-semantic conditioning. Non-semantic conditioning keeps ControlNet
active with checkerboards, stripes and other patterns that encode no text or
object-contour target. This tests whether geometric structure elicits false
reports. Positive accuracy requires the embedded target; negative content
reports count as false positives (Appendices~\ref{app:prompts},
\ref{app:negatives} and~\ref{app:fieldstats}).

\begin{table}[!t]
\centering
\small
\renewcommand{\arraystretch}{1.2}
\caption{Target recognition and false-positive reporting (\%) under the same none-capable prompt. Arrows indicate the preferred direction. Positives are contours at $s=1.0$; ControlTrace uses one checkpoint. AVR selects one answer by plurality voting.}
\label{tab:specificity_main}
\begin{tabular}{lrrrrr}
\toprule
\noalign{\vskip 3.5842pt}
Image set & $n$ & SemVink & SMSP & AVR & \textbf{ControlTrace (Ours)} \\
\noalign{\vskip 3.5842pt}
\midrule
\multicolumn{6}{l}{\textbf{Positive images: target accuracy $\uparrow$}} \\
Hidden contours & $1{,}000$ & 0.70 & 3.90 & 0.10 & 36.30 \\
\midrule
\multicolumn{6}{l}{\textbf{Negative images: false-positive rate $\downarrow$}} \\
Matched, control off & $1{,}000$ & 2.30 & 0.40 & 0.00 & 3.80 \\
Dense-texture scenes & $1{,}000$ & 1.30 & 2.00 & 0.40 & 5.20 \\
COCO photographs & $10{,}000$ & 10.54 & 9.04 & 1.45 & 5.28 \\
Non-semantic conditioning & $960$ & 2.81 & 1.15 & 0.00 & 2.92 \\
\bottomrule
\end{tabular}
\end{table}

\method{} recognized $36.3\%$ of targets, versus $0.7\%$ for SemVink,
$3.9\%$ for SMSP and $0.1\%$ for AVR. Its $2.9$--$5.3\%$ false-positive
rates (Table~\ref{tab:specificity_main}) were lower than SemVink and SMSP
on COCO but higher on other negatives. Recognition and rejection therefore
require joint assessment.

\subsection{Inference Efficiency}
\label{sec:latency}

\method{} required $146.0$\,ms per carrier, versus $138.6$\,ms for direct VLM reading, a net increase of $7.4$\,ms ($5.3\%$; Table~\ref{tab:latency}). Recovery took $12.6$\,ms including resizing, transfers and postprocessing; shorter generated answers partly offset this cost. \method{} used $384$ input tokens on average, identical to direct reading, and $72.4\%$ and $85.7\%$ fewer total tokens than SMSP and AVR, respectively. SemVink had the lowest latency and token use with its $40$-pixel input. \method{} retained one VLM call without increasing its input length, while reducing both latency and token use relative to the evaluated multi-view pipelines.

\begin{table}[!t]
\centering
\small
\setlength{\tabcolsep}{5pt}
\caption{Mean latency and VLM token use per carrier over $300$ images and two repetitions. Latency includes preprocessing and answer generation. Input counts include visual tokens; total tokens sum inputs and outputs. AVR sums all seven calls.}
\label{tab:latency}
\begin{tabular}{lrrrr}
\toprule
Method & Time (ms) $\downarrow$ & Input tokens $\downarrow$ & Output tokens $\downarrow$ & Total tokens $\downarrow$ \\
\midrule
Direct VLM & 138.6 & 384 & 2.97 & 386.97 \\
SemVink & 83.5 & 85 & 2.72 & 87.72 \\
SMSP & 373.8 & 1400 & 2.70 & 1402.70 \\
AVR & 1125.8 & 2688 & 20.01 & 2708.01 \\
\midrule
\textbf{\method{} (Ours)} & 146.0 & 384 & 2.68 & 386.68 \\
\bottomrule
\end{tabular}
\end{table}

\section{Limitations}
\label{sec:limitations}

\method{}'s advantage is concentrated in contours; SMSP performs better on strongly conditioned words, and transfer depends on the control encoder. Paired carrier--field data are required; multi-generator training and adaptation remain untested. Enhanced-scene supervision shares losses developed for control fields.

Our AVR evaluation covers fixed seven-view queries and plurality selection, excluding learned retrieval and parameter optimization. Transformations and learned configurations were selected by downstream accuracy without a held-out split; intervals do not capture selection uncertainty. PGD reveals recovery-network vulnerability, while post-attack human legibility and adversarial training remain untested. Specificity results do not estimate precision at unknown deployment prevalence.

\section{Conclusion}

On FreqBlind, recovering generator control fields improves contour recognition with a small measured increase in complete inference latency. Fidelity measurements, parameter sweeps and anagram controls distinguish structural recovery from transformed-image readability. The advantage transfers across VLM readers, while dependence on control encoders and susceptibility to adapted attacks bound its applicability.

\label{page:main_end}
\clearpage
\renewcommand{\bottomfraction}{0.3}
\label{page:ai_use}
\subsection*{AI use statement}

We used generative AI tools to assist with language polishing of the manuscript, including improvements to grammar, wording, and clarity. The authors take full responsibility for the final content.

\label{page:ethics}
\subsection*{Ethics statement}

This work builds a tool that reads content deliberately hidden inside images.
The motivating application is content moderation: prior work has shown that
hateful messages embedded this way pass both automated classifiers and
vision--language models \citep{qu2025hate}. The same tool could be used
to read hidden content that its author intended to be private, and publishing
the method and weights makes both uses easier. We judge the disclosure
favorable because the construction being defended against is already public,
cheap and demonstrated to evade deployed moderation; our experiments
evaluate image-only recovery methods against this construction.
Our experiments do not establish precision at deployment prevalence
(\S\ref{sec:specificity}); recovered content therefore requires independent
verification. The adapted attack also shows that releasing recovery weights
enables white-box evasion (\S\ref{sec:limitations}). Our benchmark contains
neutral objects and words only; it includes no hateful, personal or otherwise
sensitive content. The COCO negative set is used under its original license.

\label{page:reproducibility}
\subsection*{Reproducibility statement}

Training and evaluation code, three trained checkpoints, and FreqBlind and
training manifests are available at
\url{https://anonymous.4open.science/r/ControlTrace}.
The repository also provides raw records and analysis scripts for the main
recognition, recovery-quality and efficiency results. It documents external
model dependencies and image preparation; complete image datasets are not
included. Generation manifests record
scene prompts, targets, conditioning strengths and seeds. We retain the
evaluated images and their hashes because identical seeds do not guarantee
bitwise reproduction across software and hardware environments. The sampled
image plates also record their draw seeds.

Model, data and evaluation details are specified in \S\ref{sec:setting},
\S\ref{sec:method} and \S\ref{sec:experiments}, with reader prompts in
Appendix~\ref{app:prompts}. Trained-method results in the main recognition and recovery-quality
comparisons average three training seeds, with seed standard deviations reported
separately from item-bootstrap intervals. AVR uses fixed greedy decoding
and no training seeds. Its archive includes the frozen view configuration,
input manifests, all seven raw answers, voting diagnostics and scoring code
(Appendix~\ref{app:avr}).

\label{page:references_start}
\bibliographystyle{iclr2027_conference}
\bibliography{controltrace_references}

\appendix
\renewcommand{\floatpagefraction}{0.85}
\renewcommand{\bottomfraction}{0.7}
\setcounter{topnumber}{3}
\setcounter{bottomnumber}{2}
\setcounter{totalnumber}{4}

\section{Evaluation Protocols}
\label{app:protocols}

This appendix details the reader prompts, baseline implementations, and scoring checks introduced in Section~\ref{sec:setup}.

\subsection{Generation Settings}
\label{app:generation}

FreqBlind uses the FP16, non-EMA Realistic Vision V5.1 checkpoint and QR
Code Monster for SD1.5. Carriers are $512\times512$ pixels, generated in
FP16 with the PNDM scheduler, $30$ inference steps and classifier-free
guidance scale $7.5$. The retained manifest specifies the prompt, negative
prompt, control image, conditioning strength and seed for each carrier.

The training pairs instead use the standard SD1.5 base checkpoint with
the same control encoder and $20$ inference steps. Thus, the two datasets
share a generator architecture and control encoder, but not identical
base weights or sampling budgets. The training-generation records specify
PyTorch $2.3.1$ with CUDA $12.1$ and Diffusers $0.35.2$.

The $50$ contour controls comprise $20$ pre-existing animal images and
$30$ object shapes rendered from SVG files. The latter are centered on
a $512\times512$ canvas, targeting $20\%$ foreground coverage with a
longest-side cap of $92\%$ of the canvas. Controls are converted to
grayscale and inverted to bright structure on a dark background. Words
use Lato Black with zero stroke expansion and $-10$-pixel letter spacing;
the renderer fits each string within $86\%$ of the canvas width and $30\%$
of its height. The control inventory records the exact source files.

Scene assignment uses seed $20260908$ and excludes prompts containing
the target name or its listed aliases before drawing $20$ scenes per
target. A separate seed is recorded for each target--scene--strength
record. The two domains reuse the same seed sequence; seeds are not
unique across the full benchmark. Checkpoint hashes identify the retained
local artifacts; upstream commit revisions were not recorded.

\subsection{Reader Prompts}
\label{app:prompts}

The FreqBlind contour task prompt used for SemVink, SMSP and AVR is:

\begin{quote}\small
The image contains a hidden ANIMAL or OBJECT pattern formed by the arrangement
of scene elements. Identify what animal or object is hidden. Answer with ONLY
the name (e.g., 'panda', 'cat'), nothing else.
\end{quote}

The word-domain prompt, used for both real words and pronounceable non-words, is:

\begin{quote}\small
The image contains a hidden WORD formed by the arrangement of objects,
structures, or visual patterns. Identify the hidden word. Answer with ONLY
the word, nothing else.
\end{quote}

All four methods use the following complete prompt for positive and
negative specificity inputs, without additional view-specific text:

\begin{quote}\small
The image contains a hidden ANIMAL or OBJECT pattern formed by the arrangement
of scene elements. Identify what animal or object is hidden. Answer with ONLY
the name (for example, 'panda' or 'cat'), nothing else. If there is no hidden
animal or object, answer with ONLY 'none'.
\end{quote}

For FreqBlind recognition, the SemVink task prompt is preceded by:

\begin{quote}\small
This image has been converted to a low-resolution global view. Focus on
global shape rather than local texture.
\end{quote}

For FreqBlind recognition, the SMSP task prompt uses the following prefix:

\begin{quote}\small
I provide four views of the SAME image: the original and three low-frequency
global views. There is hidden content embedded in these images. With the help
of the views, what is it?
\end{quote}

These prefixes apply to recognition; specificity uses the shared prompt alone.
AVR asks the task question independently for each view,
without a view-description prefix. In the three-seed main ControlTrace
evaluation, the contour prompt uses ``for example, 'panda' or 'cat''' in
place of ``e.g., 'panda', 'cat'''. Generation limits are specified in
Appendix~\ref{app:avr}.

The alternative HC-Bench regime uses the questions quoted by SemVink
\citep{li2025semvink}. For hidden objects:

\begin{quote}\small
What is within this image? Is there any other content hidden within this image?
\end{quote}

For hidden text:

\begin{quote}\small
What is within this image? Is there any text hidden within this image?
\end{quote}

HC-Bench's other regime uses the FreqBlind task prompts above. Neither
regime supplies the correct target, a candidate list or an answer-informed
follow-up. AVR's seven views use the same question within each regime.

\subsection{Baseline Implementations}
\label{app:avr}

\paragraph{SemVink.}
SemVink resizes the RGB carrier to a width of $40$ pixels using bilinear
interpolation, preserving its aspect ratio. The reader receives this view
with the task prompt and, for recognition, the global-shape prefix given above.

\paragraph{SMSP.}
Our SMSP implementation converts the carrier to grayscale and applies
hard circular masks around the centered Fourier origin. The radius ratios
are $0.012$, $\sqrt{0.012\times0.05}$ and $0.05$, relative to the shorter
image side; integer truncation gives radii $6$, $12$ and $25$ for
$512\times512$ inputs. Each inverse transform is converted to its
magnitude, min--max scaled to $[0,255]$ and stored as an RGB image.

The filtered images are resized with Lanczos interpolation to $100$, $200$
and $400$ pixels square, respectively. Each is centered on a white canvas
with the original carrier's dimensions. The reader receives the original
RGB carrier followed by these three views in increasing size order and
returns one answer. Recognition uses the four-view prefix given above;
specificity uses the shared none-capable prompt without that prefix.

\paragraph{AVR.}
AVR Experiment 3 queries seven transformed views independently
\citep{chen2026now}. We implement this VLM protocol without the
template bank, learned CLIP projection, query gate or calibration head.
The published processing sequences do not specify numerical parameters;
Table~\ref{tab:avrviews} gives the values fixed before recognition
evaluation. These values instantiate the described operations and are not
claimed to reproduce the authors' settings or their optimum.

\begin{table}[!htbp]
\caption{Fixed parameters of our AVR seven-view implementation, in evaluation
order. Values refer to a longest image side of $512$ pixels. Gaussian and
box values are Pillow radius arguments. Each output retains the carrier's
dimensions before the reader's usual preprocessing.}
\label{tab:avrviews}
\begin{center}
\small
\begin{tabular}{p{0.16\linewidth}p{0.76\linewidth}}
\toprule
View & Operations and parameters \\
\midrule
Original & RGB copy. \\
Contrast & Gaussian radius $1$; YCbCr conversion; equalize Y only; RGB conversion. \\
Closure-edge & Grayscale; $3\!\times\!3$ Sobel magnitude; min--max normalization; $3\!\times\!3$ closing once; Gaussian radius $1$; normalization. \\
Mask & Grayscale; Gaussian radius $4$; normalization; $256$-bin Otsu threshold; $3\!\times\!3$ opening and closing, once each. \\
Inverse-mask & $255-M$, using the cleaned mask $M$ above. \\
Foreground & Original RGB foreground selected by $M$ over RGB $(128,128,128)$. \\
Low-pass & Gaussian radius $16$; box radius $8$; median $5\!\times\!5$; grayscale; unsharp Gaussian radius $2$, amount $0.5$; normalization. \\
\bottomrule
\end{tabular}
\end{center}
\end{table}

For other sizes, Gaussian and box radii scale by
$\rho=\max(W,H)/512$. Morphological and median radii scale by $\rho$, round
half up and yield odd widths capped by the smaller dimension. Sobel retains
its $3\times3$ kernel, uses replicated boundaries and takes the Euclidean
gradient magnitude. Min--max normalization uses nearest-even rounding;
constant maps retain their level. Otsu chooses the smallest tied threshold,
marks values strictly above it as foreground and returns all background
when no nondegenerate split exists. Sharpening computes
$\operatorname{clip}(G+0.5(G-\operatorname{Gaussian}_{2\rho}(G)),0,255)$
before normalization. The release records the exact Pillow/NumPy code and
configuration hashes.

\paragraph{Single-answer selection and any-view coverage.}
Our plurality adapter groups complete answers after Unicode NFKC
normalization, case folding and whitespace collapse. It retains punctuation
and uses no target labels or aliases. The most frequent answer wins; a tie
selects the earliest supporting view in the table's order. We score that
view's original answer using the experiment's recognition rule. Separately,
any-view coverage counts an image as correct when at least one of its seven
answers is correct, following AVR Experiment 3. This reference-based
coverage does not provide an inference-time selector. Complete seven-view
diagnostics are retained in the accompanying source data.

Empty responses and responses without a Unicode letter or numeral are
invalid. They remain in voting and in the evaluation denominator; a
selected invalid response is an error and never a correct rejection.
The main contour evaluation selects the original view in $58.8\%$ of
cases and has tied winning answers in $26.3\%$, illustrating the limits
of exact-answer voting when views elicit different descriptions.

\paragraph{Specificity.}
All four methods use the same none-capable task prompt on positive and
negative images, without additional view-specific prefixes.
It retains the assertion that hidden content is present while allowing an
explicit ``none'' response. Reported rates are conditional on this wording,
not upper bounds for other prompts.
For all four methods, after separating invalid outputs, the primary
comparison retains the existing rejection parser: ASCII normalization followed by an initial
``none'', ``nothing'' or ``no''; valid responses with no
remaining ASCII tokens also count as rejections. A stricter diagnostic
accepts only the complete answers ``none'', ``nothing'',
``no'', ``no hidden shape'' or ``nothing hidden'', after vote
normalization and removal of one final ASCII period. These parsers score
each method's final answer. AVR's all-seven-NONE and any-view-content
rates remain separate diagnostics. Any-NONE is not counted as correct rejection.

\paragraph{Inference settings.}
AVR uses batch size one, deterministic decoding and seven independent
reader calls per image. The generation limits are $128$ tokens per view
for FreqBlind recognition, $24$ for specificity and $160$ for HC-Bench.
The SemVink and SMSP specificity evaluations use batch size one,
deterministic decoding and a $24$-token limit, with one reader call per image.
SMSP supplies all four views jointly within that single reader call.
The main ControlTrace evaluation uses a $64$-token limit. These limits
specify the permitted output length, rather than measured token use.

The separate efficiency experiment used a common $128$-token limit for all methods (\S\ref{sec:setup}). Token counts came from the actual sequences passed to and returned by the reader, including chat-format and generated ending tokens. We summed all calls per carrier, then averaged repetitions and carriers; no call reached the generation limit. Token counts measure sequence use, not monetary charges.

The $1{,}800$-image Qwen7B reader subset uses the corresponding main
recognition outputs under the same inference conditions. Image
perturbations are applied to the carrier before view construction.
PGD optimizes against ControlTrace, so AVR scores on these carriers
measure attack transfer. For recovery-quality evaluation,
Appendix~\ref{app:quality} reports each transformed view and their
unweighted mean without reference-based view selection.

\subsection{Scoring Sensitivity}
\label{app:judge}

FreqBlind uses the substring-or-$80\%$-of-target-tokens rule in
Section~\ref{sec:setup}. Table~\ref{tab:judge} compares it with an
alternative rule requiring the target's final token to appear as a complete
token in the response. This changes both substring and token matching, so
it is not uniformly more permissive. AVR's selected answer is fixed before
either scorer is applied. Results are aggregated across strengths, and \method{} uses
seed $42$ here, rather than the three-seed mean of Table~\ref{tab:main}.

\begin{table}[!htbp]
\centering
\small
\caption{Recognition accuracy (\%) under the reported scoring rule and an alternative requiring the final target token to appear as a complete response token. Results aggregate the three conditioning strengths. ControlTrace uses seed $42$; AVR uses the same plurality-selected answer under both rules. Higher values indicate better recognition.}
\label{tab:judge}
\begin{tabular}{lrrrr}
\toprule
& \multicolumn{2}{c}{Contours $\uparrow$} & \multicolumn{2}{c}{Words $\uparrow$} \\
\cmidrule(lr){2-3}\cmidrule(lr){4-5}
Method & Reported & Alternative & Reported & Alternative \\
\midrule
Carrier, read directly & 6.30 & 6.40 & 2.57 & 2.40 \\
SemVink & 19.43 & 19.77 & 53.60 & 51.33 \\
SMSP & 33.33 & 34.17 & 63.37 & 62.90 \\
AVR & 19.37 & 19.77 & 23.03 & 22.37 \\
\midrule
\textbf{ControlTrace (Ours)} & 60.10 & 62.00 & 61.23 & 59.90 \\
\bottomrule
\end{tabular}
\end{table}

Changing the response-matching rule can raise or lower a score. In
particular, complete-token matching excludes embedded substrings, while
matching only the final target token can accept less specific responses.
These results assess sensitivity to automatic judgement without changing
the images, reader outputs or AVR selection rule.

\section{Recognition Performance}
\label{app:extended}

This appendix expands Section~\ref{sec:main} with intermediate conditioning strengths, HC-Bench implementation checks, and paired lexical controls.

\subsection{Intermediate Conditioning Strengths}
\label{app:band}

\begin{figure}[H]
\begin{center}
\includegraphics[width=\linewidth]{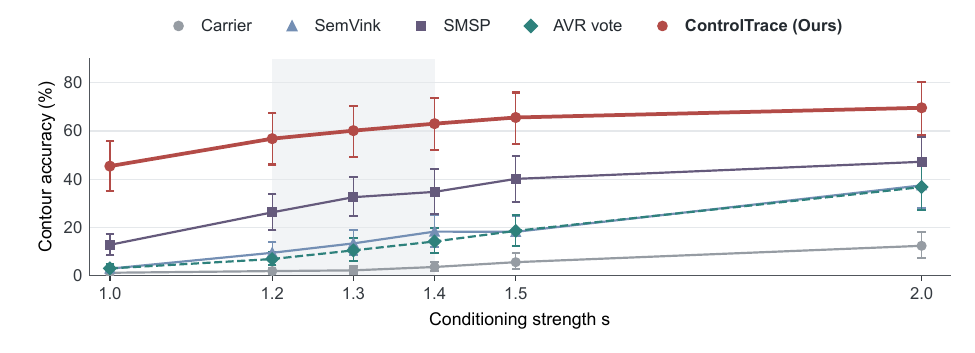}
\end{center}
\caption{Single-answer contour recognition across conditioning strengths. Each strength contains $1{,}000$ carriers; ControlTrace averages three training seeds. Error bars show $95\%$ item-bootstrap intervals from $20{,}000$ resamples of $50$ targets. AVR uses its fixed plurality answer. The shaded region marks the additional strengths $s=1.2$--$1.4$.}

\label{fig:band}
\end{figure}

We evaluate $3{,}000$ additional contour carriers at
$s \in \{1.2, 1.3, 1.4\}$ (Figure~\ref{fig:band}), covering the
conditioning range used in HC-Bench \citep{li2025semvink}. These carriers
share FreqBlind's $50$ targets and $20$ scenes per target, with an
independent generation seed for each record. All methods use the
recognition protocol in Appendix~\ref{app:protocols}.

The contour-recognition advantage extends across the added strengths.
\method{} reaches $56.8\%$, $60.1\%$ and $63.0\%$, compared with
$9.4\%$, $13.3\%$ and $18.2\%$ for SemVink, $26.2\%$, $32.5\%$
and $34.7\%$ for SMSP, and $6.8\%$, $10.4\%$ and $14.1\%$ for
AVR plurality. AVR any-view coverage is $32.5\%$, $39.6\%$ and $44.8\%$.
These results do not isolate the causes of the difference from SemVink's
published performance; the HC-Bench evaluation below examines dataset,
prompt and scoring differences.

SemVink reported $91$--$100\%$ accuracy on HC-Bench under a different dataset, reader, and judgement protocol. That published range provides context for the implementation check and is not plotted as a directly comparable FreqBlind reference.

\subsection{HC-Bench Evaluation}
\label{app:hcbench}

We compare the baseline implementations on HC-Bench's $56$ hidden-object
and $28$ Latin-text images using Qwen2.5-VL-7B-Instruct
(Table~\ref{tab:hcbench}). Each method uses the same transformation
implementation as on FreqBlind and is evaluated with both SemVink's
quoted question and the FreqBlind task question. Scoring follows the
FreqBlind matching rule with HC-Bench's word-spacing aliases, including
``newyork''/``new york'' and ``notredame''/``notre dame''.
The scorer is shared across methods and is not used to select AVR's answer.

The published AVR evaluation instead uses all $112$ HC-Bench images,
including Chinese text, with Qwen2-VL-2B and any-view scoring. Our
$84$-image subset, reader and automatic scoring protocol define a
separate comparison.

\begin{table}[!htbp]
\centering
\small
\setlength{\tabcolsep}{5pt}
\caption{HC-Bench baseline recognition accuracy (\%) under two prompt regimes. The shared automatic substring/token rule includes word-spacing aliases. Each regime uses the same $56$ object and $28$ Latin-text images. Downscaling rows give the output width; a dash marks an unevaluated setting. Higher values indicate better recognition.}
\label{tab:hcbench}
\begin{tabular}{lrrrr}
\toprule
& \multicolumn{2}{c}{SemVink quoted question} & \multicolumn{2}{c}{FreqBlind task question} \\
\cmidrule(lr){2-3}\cmidrule(lr){4-5}
Input / method & Objects $\uparrow$ & Text $\uparrow$ & Objects $\uparrow$ & Text $\uparrow$ \\
\midrule
Carrier, read directly & 3.6 & 3.6 & --- & --- \\
Downscale to $32$\,px & 21.4 & 28.6 & 21.4 & 25.0 \\
Downscale to $40$\,px & 28.6 & 35.7 & 30.4 & 28.6 \\
Downscale to $64$\,px & 28.6 & 50.0 & 21.4 & 50.0 \\
Downscale to $85$\,px & 26.8 & 57.1 & 25.0 & 53.6 \\
Downscale to $128$\,px & 17.9 & 46.4 & 14.3 & 50.0 \\
SMSP & 19.6 & 10.7 & 19.6 & 42.9 \\
AVR & 3.6 & 3.6 & 8.9 & 21.4 \\
\bottomrule
\end{tabular}
\end{table}

Replacing SemVink's quoted question with the FreqBlind task question
changes each downsampling score by at most $7.2$ points and raises
SMSP's text score from $10.7\%$ to $42.9\%$. AVR uses the same view
bank and voting rule under both questions. Its single-answer scores
increase in both domains (Table~\ref{tab:hcbench}). The effect of the
prompt therefore depends on the target domain and method.

HC-Bench also differs from FreqBlind in what a correct name requires.
Its object subset comprises $36$ common nouns, six person or scene
categories, and $14$ named entities or non-English labels. These targets
can require more specific names than FreqBlind's common-noun contours;
a general category answer may therefore fail the automatic matching rule.

SemVink improves over direct reading on HC-Bench but remains below its
published result. Differences in targets, readers, prompts and scoring
prevent attributing the gap to a single cause. The matched-input and
matched-reader comparisons on FreqBlind therefore evaluate the stated
implementations, rather than reproductions of each method's published
headline score.

\subsection{Lexical Controls}
\label{app:lexicalanalysis}

\begin{figure}[!htbp]
\begin{center}
\includegraphics[width=\linewidth]{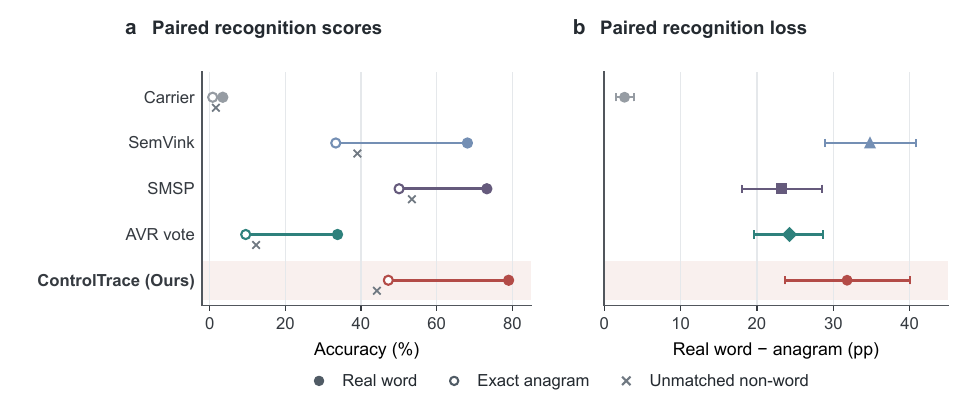}
\end{center}
\caption{Real-word and non-word recognition averaged across conditioning strengths. Paired endpoints compare $25$ real words with exact anagrams using the same letter multiset, font, and rendering procedure ($1{,}500$ carriers per target type). Unmatched benchmark non-words are shown separately. Difference intervals are $95\%$ paired bootstraps over the $25$ word pairs; ControlTrace averages three training seeds. AVR uses its plurality-selected answer; seven-answer coverage is retained in the source data.}

\label{fig:lexical}
\end{figure}

The benchmark word domain contains real words and unmatched pronounceable non-words. A separate paired experiment replaced each real word with an exact anagram, preserving its letters and rendering procedure while changing their order and spatial arrangement. Figure~\ref{fig:lexical} separates these paired comparisons from the unmatched benchmark controls.

Real-word advantages appeared for every evaluated method. The aggregate paired drops in single-answer accuracy were $34.8$ points for SemVink, $23.2$ for SMSP, $24.3$ for AVR plurality ($33.8\%$ to $9.5\%$), and $31.8$ for ControlTrace. AVR any-view coverage decreased from $57.3\%$ to $26.8\%$. These differences support a lexical contribution, but the simultaneous change in spatial arrangement prevents attributing the full effect to lexical completion.

\section{Recovery Quality and Training Design}
\label{app:training}

This appendix provides implementation details for Section~\ref{sec:method}, training-data details for Section~\ref{sec:setup}, and reconstruction-quality evaluation and training controls for Section~\ref{sec:nafnet}.

\net{} takes a $256\times256$ RGB carrier without a generation prompt or
text encoder. The four encoder stages have $48$, $96$, $192$ and $384$
channels, followed by a $384$-channel bottleneck at $16\times16$ resolution.
Each block contains two $3\times3$ convolutions, each followed by group
normalization with eight groups and a SiLU activation. Downsampling uses
$2\times2$ average pooling. The decoder uses bilinear upsampling and
concatenated skip connections, with $192$, $192$, $96$ and $48$ output
channels from coarse to fine. The grayscale and edge heads use $1\times1$
convolutions; the grayscale output uses a sigmoid. Table~\ref{tab:implementation}
summarizes the reference training configuration.

\begin{table}[!htbp]
\centering
\small
\caption{Implementation settings for the reference ControlTrace model. Training batches group six carrier variants from each of four control images.}
\label{tab:implementation}
\begin{tabular}{ll}
\toprule
Setting & Value \\
\midrule
Optimizer & AdamW \\
Learning rate & $2\times10^{-4}$ \\
Weight decay & $10^{-4}$ \\
Training batch size & $24$ ($4$ controls $\times$ $6$ variants) \\
Validation batch size & $16$ \\
Training duration & $30$ epochs \\
Training seeds & $42$, $43$, $44$ \\
Checkpoint selection & Minimum validation loss \\
\bottomrule
\end{tabular}
\end{table}

Equation~\ref{eq:training_loss} defines the training objective. Edge
cross-entropy uses logits with no pixel or class weighting.
The cosine term uses flattened fields and
$d(x)=\max\{\|x\|_2,\varepsilon_c\}$. The Sobel operation uses convolution
$\ast$ with one-pixel zero padding; its squared magnitude is floored before
taking the square root. We set $\varepsilon_c=10^{-8}$ and
$\varepsilon_s=10^{-6}$:

\begin{equation}
\begin{aligned}
\cos_{\varepsilon}(\hat{g},c)
 &= \frac{\hat{g}^{\mathsf T}c}{d(\hat{g})d(c)},\\
E(x) &= \min\!\left(1,\sqrt{\max\!\left(
 (K_x\ast x)^2+(K_y\ast x)^2,\varepsilon_s\right)}\right),\\
K_x &= \begin{bmatrix}-1&0&1\\-2&0&2\\-1&0&1\end{bmatrix},
\qquad K_y=K_x^{\mathsf T}.
\end{aligned}
\end{equation}

\subsection{Control-Field Fidelity}
\label{app:quality}

We evaluate all $6{,}000$ FreqBlind carriers against the control images
used to generate them, retaining $1{,}000$ images per domain--strength
cell. Reference images are converted to grayscale, resized with bicubic
interpolation to $256\times256$, and divided by $255$. ControlTrace and
NAFNet use their three existing checkpoints, with unquantized floating-point
predictions clipped to $[0,1]$. No model is retrained for this evaluation.

Transformation views use the implementations and settings from the recognition
experiments. Each view is converted to grayscale, resized with bicubic
interpolation to the same reference grid, and divided by $255$.
For SMSP, we first remove the known canvas padding around its $100$, $200$
and $400$-pixel content regions, then resize those regions. The original
SMSP view remains unchanged before common-grid conversion. This deterministic
mapping reverses view placement; it does not search for alignment using
the reference. AVR retains all seven fixed views, and direct Gaussian
blur uses $\sigma=16$. We apply no additional intensity normalization,
polarity inversion, or reference-dependent view selection.

For an evaluated field $x$ and reference $c$ with $N$ pixels,
the pixel-error metrics are
\begin{equation}
\operatorname{MSE}(x,c)=\frac{1}{N}\sum_{i=1}^{N}(x_i-c_i)^2,
\qquad
\operatorname{MAE}(x,c)=\frac{1}{N}\sum_{i=1}^{N}|x_i-c_i|.
\end{equation}
SSIM compares corresponding local windows using their Gaussian-weighted
means, variances and covariance \citep{wang2004image}:
\begin{equation}
\operatorname{SSIM}(x,c)=\frac{1}{|\Omega|}\sum_{w\in\Omega}
\frac{(2\mu_{x,w}\mu_{c,w}+C_1)(2\sigma_{xc,w}+C_2)}
{(\mu_{x,w}^2+\mu_{c,w}^2+C_1)(\sigma_{x,w}^2+\sigma_{c,w}^2+C_2)}.
\end{equation}
We use an $11\times11$ Gaussian window with standard deviation $1.5$,
population covariance, $C_1=0.01^2$ and $C_2=0.03^2$. The set $\Omega$
excludes five boundary pixels on each side. Lower MSE and MAE indicate
smaller pixel errors; higher SSIM indicates greater local structural similarity.

Metrics are calculated separately for each output. SMSP and AVR summaries
average their four and seven view scores per image, respectively;
trained-model summaries average scores from three seeds per image.
These operations average scores, not images. Dataset summaries then
average images, giving equal weight to the three conditioning strengths.
View means describe the fixed view banks, not a fused reconstruction
or the quality of an answer selected by the VLM. Per-view results are
reported separately to retain variation within each bank.

Uncertainty is estimated by resampling the $50$ targets in each domain
$20{,}000$ times, keeping all scenes, strengths, views and seeds of each
target together. Seed variability is recorded separately from item-bootstrap
uncertainty. The metrics measure fidelity to the generating field;
they do not replace open-ended recognition, and are not the training
objectives of the transformation-based methods.

As a diagnostic, an all-zero field obtained MSE/MAE/SSIM of
$0.2090/0.2166/0.6445$ on contours and $0.1149/0.1185/0.7870$ on words.
Large shared backgrounds can yield high SSIM without recovered content.
\method{} improved all three metrics over this reference in both domains;
SSIM should nevertheless be interpreted alongside pixel errors and recognition.

Tables~\ref{tab:quality_strength} and~\ref{tab:quality_views} report
strength-specific and individual-view means. Per-image scores, bootstrap
intervals and seed summaries accompany the evaluation scripts.

\begin{table}[!htbp]
\centering
\small
\setlength{\tabcolsep}{4pt}
\caption{Control-field fidelity by conditioning strength, with $1{,}000$ carriers per domain--strength cell. The view and seed aggregation rules match Table~\ref{tab:quality_main}; values are means.}
\label{tab:quality_strength}
\begin{tabular}{llrrrrrr}
\toprule
& & \multicolumn{3}{c}{Contour} & \multicolumn{3}{c}{Word} \\
\cmidrule(lr){3-5}\cmidrule(lr){6-8}
Method & $s$ & MSE $\downarrow$ & MAE $\downarrow$ & SSIM $\uparrow$ & MSE $\downarrow$ & MAE $\downarrow$ & SSIM $\uparrow$ \\
\midrule
Carrier & 1.0 & 0.1381 & 0.3201 & 0.0855 & 0.1491 & 0.3319 & 0.0453 \\
 & 1.5 & 0.0990 & 0.2651 & 0.1322 & 0.1161 & 0.2897 & 0.0650 \\
 & 2.0 & 0.0777 & 0.2343 & 0.1600 & 0.0979 & 0.2683 & 0.0760 \\
\midrule
SemVink & 1.0 & 0.1327 & 0.3270 & 0.1115 & 0.1421 & 0.3361 & 0.0546 \\
 & 1.5 & 0.0980 & 0.2786 & 0.1325 & 0.1119 & 0.2977 & 0.0659 \\
 & 2.0 & 0.0805 & 0.2522 & 0.1388 & 0.0963 & 0.2787 & 0.0688 \\
\midrule
SMSP (view mean) & 1.0 & 0.1490 & 0.3389 & 0.0998 & 0.1699 & 0.3599 & 0.0491 \\
 & 1.5 & 0.1111 & 0.2883 & 0.1231 & 0.1334 & 0.3163 & 0.0606 \\
 & 2.0 & 0.0916 & 0.2609 & 0.1349 & 0.1171 & 0.2980 & 0.0653 \\
\midrule
AVR (view mean) & 1.0 & 0.2708 & 0.4001 & 0.1415 & 0.2793 & 0.4070 & 0.1187 \\
 & 1.5 & 0.2555 & 0.3804 & 0.1744 & 0.2654 & 0.3907 & 0.1352 \\
 & 2.0 & 0.2474 & 0.3691 & 0.1971 & 0.2587 & 0.3829 & 0.1500 \\
\midrule
Gaussian blur ($\sigma=16$) & 1.0 & 0.1471 & 0.3527 & 0.1054 & 0.1478 & 0.3504 & 0.0457 \\
 & 1.5 & 0.1210 & 0.3185 & 0.1115 & 0.1240 & 0.3204 & 0.0492 \\
 & 2.0 & 0.1076 & 0.2990 & 0.1120 & 0.1117 & 0.3055 & 0.0488 \\
\midrule
NAFNet & 1.0 & 0.0593 & 0.0816 & 0.7486 & 0.0566 & 0.0831 & 0.7367 \\
 & 1.5 & 0.0251 & 0.0399 & 0.8503 & 0.0291 & 0.0459 & 0.8267 \\
 & 2.0 & 0.0178 & 0.0310 & 0.8789 & 0.0239 & 0.0387 & 0.8484 \\
\midrule
\textbf{\method{} (Ours)} & 1.0 & 0.0561 & 0.0782 & 0.7601 & 0.0509 & 0.0755 & 0.7645 \\
 & 1.5 & 0.0242 & 0.0393 & 0.8533 & 0.0244 & 0.0391 & 0.8493 \\
 & 2.0 & 0.0178 & 0.0312 & 0.8802 & 0.0205 & 0.0333 & 0.8679 \\
\bottomrule
\end{tabular}
\end{table}

\begin{table}[!htbp]
\centering
\small
\setlength{\tabcolsep}{4pt}
\caption{Fidelity of individual SMSP and AVR views, averaged over $3{,}000$ carriers per domain. These scores retain variation within each view bank; no view is selected using the control field.}
\label{tab:quality_views}
\begin{tabular}{llrrrrrr}
\toprule
& & \multicolumn{3}{c}{Contour} & \multicolumn{3}{c}{Word} \\
\cmidrule(lr){3-5}\cmidrule(lr){6-8}
Method & View & MSE $\downarrow$ & MAE $\downarrow$ & SSIM $\uparrow$ & MSE $\downarrow$ & MAE $\downarrow$ & SSIM $\uparrow$ \\
\midrule
SMSP & Original & 0.1049 & 0.2732 & 0.1259 & 0.1210 & 0.2966 & 0.0621 \\
 & Low-pass, 100 px & 0.1557 & 0.3485 & 0.1048 & 0.1920 & 0.3872 & 0.0451 \\
 & Low-pass, 200 px & 0.1068 & 0.2867 & 0.1184 & 0.1289 & 0.3158 & 0.0607 \\
 & Low-pass, 400 px & 0.1015 & 0.2757 & 0.1279 & 0.1186 & 0.2994 & 0.0654 \\
\midrule
AVR & Original & 0.1049 & 0.2732 & 0.1259 & 0.1210 & 0.2966 & 0.0621 \\
 & Contrast & 0.1725 & 0.3387 & 0.1369 & 0.2338 & 0.4018 & 0.0697 \\
 & Closure edge & 0.1764 & 0.2699 & 0.0633 & 0.1027 & 0.1920 & 0.0729 \\
 & Mask & 0.1614 & 0.1745 & 0.6385 & 0.2720 & 0.2822 & 0.5181 \\
 & Inverse mask & 0.8093 & 0.8225 & 0.0286 & 0.7068 & 0.7170 & 0.1267 \\
 & Foreground & 0.2219 & 0.4550 & 0.0858 & 0.2370 & 0.4745 & 0.0408 \\
 & Low pass & 0.1588 & 0.3485 & 0.1182 & 0.2015 & 0.3907 & 0.0523 \\
\bottomrule
\end{tabular}
\end{table}

\subsection{Supervision and Backbone Controls}
\label{app:targets}

We compare control-field supervision with a stronger-conditioning scene
target to assess the choice of reconstruction target. The scene target uses
the same prompt and generation seed as the input carrier, regenerated at
$s=2.0$ and converted to grayscale. This comparison holds the U-Net, data,
losses, optimization budget and training seeds fixed; loss weights are not
retuned for the scene target.

To assess sensitivity to the recovery backbone, Table~\ref{tab:objective}
also includes NAFNet with approximately matched capacity and training budget.
NAFNet received no additional hyperparameter tuning, whereas the U-Net depth
had been selected in a separate sweep.

\begin{table}[!htbp]
\caption{Contour recognition for supervision-target and backbone controls.
Accuracy (\%) is averaged over three training seeds, with $1{,}000$ carriers
per conditioning strength. The All strengths column aggregates the three evaluated
strengths. Higher values indicate better recognition.}
\label{tab:objective}
\begin{center}
\small
\setlength{\tabcolsep}{7pt}
\begin{tabular}{llrr}
\toprule
Supervision target & Backbone & $s{=}1.0$ $\uparrow$ & All strengths $\uparrow$ \\
\midrule
Scene at $s{=}2.0$ & U-Net & 13.27 & 29.13 \\
Control field & NAFNet & 38.13 & 56.51 \\
\midrule
\best{Control field} & \best{U-Net (Ours)} & \best{45.40} & \best{60.21} \\
\bottomrule
\end{tabular}
\end{center}
\end{table}

At $s=1.0$, replacing the stronger-conditioning scene target with the
control field increased accuracy by $32.13$ points ($95\%$ paired
item-bootstrap CI, $[23.90,40.87]$). Under control-field supervision,
replacing NAFNet with the U-Net increased accuracy by $7.27$ points
($[4.83,9.77]$). These comparisons average three training seeds on the
same $1{,}000$ carriers.

\subsection{Training Configurations}
\label{app:factorial}

The complete pool contains $36{,}000$ carrier--control pairs across three
conditioning strengths, using $2{,}000$ FIGR-8 control images and $750$ scene
prompts. The reference model uses the $12{,}000$ pairs at $s=1.0$, with
seeds $42$, $43$ and $44$. FreqBlind instead uses separate silhouette images
for contours and rendered glyphs for words. None of its control images or
$200$ scene prompts appears in the training pool. The reference uses three-channel carrier inputs.
Table~\ref{tab:ablation_main} reports the component comparisons, averaged
over these three training seeds.

\subsection{Model Selection}
\label{app:pixel}

Within each training run, we selected the checkpoint with the lowest loss on
the internal validation split. FreqBlind uses separate control images and
scene prompts, but its recognition results informed selection among
training configurations. We did not reserve a separate subset for
configuration selection. The fixed-operation sweeps characterize parameter
sensitivity; their best observed settings were identified on the evaluated
data. Item-bootstrap intervals quantify target-item variation for the
selected configurations and do not account for the configuration-selection
process.

Table~\ref{tab:seed_paired} reports variability across three independently
trained checkpoints on the same evaluation images. Seed standard deviations
describe training variability and are distinct from the item-bootstrap
intervals used for method comparisons.

\begin{table}[!htbp]
\centering
\small
\caption{Training-seed variability of \textbf{ControlTrace (Ours)} on FreqBlind.
Each domain contains $50$ target items and $1{,}000$ carriers per strength.
The seed scores and Mean column report recognition accuracy (\%); SD is the sample
standard deviation across the three training seeds, in percentage points.
The All rows aggregate the three strengths within each seed.}
\label{tab:seed_paired}
\setlength{\tabcolsep}{5pt}
\begin{tabular}{llrrrrr}
\toprule
& & \multicolumn{4}{c}{Accuracy (\%) $\uparrow$} & \\
\cmidrule(lr){3-6}
Domain & $s$ & Seed 42 & Seed 43 & Seed 44 & Mean & SD (pp) \\
\midrule
Contours & 1.0 & 45.10 & 46.80 & 44.30 & 45.40 & 1.28 \\
 & 1.5 & 65.10 & 66.90 & 64.80 & 65.60 & 1.14 \\
 & 2.0 & 70.10 & 68.80 & 70.00 & 69.63 & 0.72 \\
 & All & 60.10 & 60.83 & 59.70 & 60.21 & 0.57 \\
\midrule
Words & 1.0 & 27.40 & 29.60 & 27.80 & 28.27 & 1.17 \\
 & 1.5 & 75.10 & 75.70 & 74.40 & 75.07 & 0.65 \\
 & 2.0 & 81.20 & 83.10 & 80.30 & 81.53 & 1.43 \\
 & All & 61.23 & 62.80 & 60.83 & 61.62 & 1.04 \\
\bottomrule
\end{tabular}
\end{table}

\begin{minipage}{\linewidth}
\section{Transformation Sensitivity}
\label{app:analysis}

The method comparison in Section~\ref{sec:main} evaluates SemVink, SMSP
and AVR as complete recognition pipelines. Here, we sweep fixed image
operations to examine how transformation parameters affect recognition.
Gaussian blur and downsampling expose the balance between suppressing
scene texture and preserving hidden structure (Figure~\ref{fig:recovery_analysis}).

\end{minipage}

\subsection{Fixed-Operation Sweep}
\label{app:sweep}

\begin{figure}[!htbp]
\centering
\includegraphics[width=\linewidth]{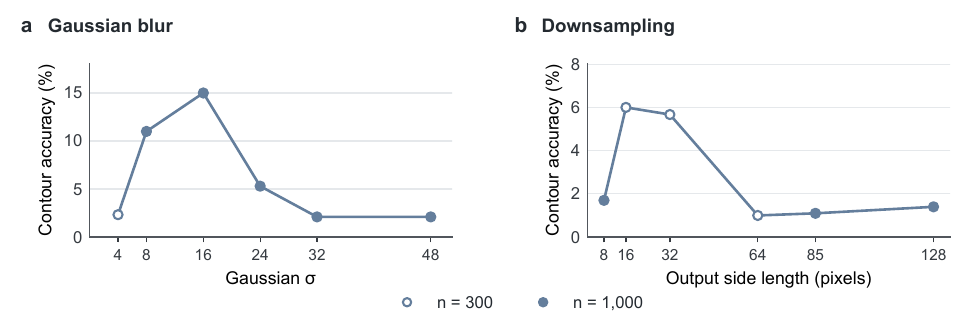}
\caption{Parameter sensitivity on weak contours. Gaussian blur and downsampling vary within the tested grids; hollow markers use $300$ carriers and filled markers use $1{,}000$. The panels use separate vertical scales. Lines connect tested settings and do not imply an exhaustive search.}
\label{fig:recovery_analysis}
\end{figure}

\begin{table}[!htbp]
\centering
\small
\caption{Selected transformation families. Aggregate cells use $3{,}000$ carriers per domain; weak-condition cells use $1{,}000$. The final column is the contour-minus-word difference in percentage points. Method comparisons appear in Table~\ref{tab:main}; Table~\ref{tab:sweep} lists all $27$ operations.}
\label{tab:naive}
\begin{tabular}{lrrrrr}
\toprule
& \multicolumn{2}{c}{Contour $\uparrow$} & \multicolumn{2}{c}{Word $\uparrow$} & \\
Operation / method & All & $s{=}1.0$ & All & $s{=}1.0$ & C$-$W \\
\midrule
Sobel (edge operator) & 8.03 & 1.00 & 4.67 & 0.00 & $+3.37$ \\
Otsu (pointwise map) \citep{otsu1979threshold} & 12.17 & 2.00 & 18.23 & 0.10 & $-6.07$ \\
blur $+$ histogram equalization \citep{qu2025hate} & 29.53 & 10.10 & 56.30 & 13.10 & $-26.77$ \\
Gaussian blur $\sigma{=}8$ & 34.83 & 11.00 & 59.83 & 13.80 & $-25.00$ \\
Gaussian blur $\sigma{=}16$ & 34.33 & 15.00 & 39.17 & 7.50 & $-4.83$ \\
\bottomrule
\end{tabular}
\end{table}

The sweep contains $20$ operations initially evaluated on $300$ carriers
per domain--strength cell and seven additional settings near the grid
boundaries. Selected initial settings and all seven additional settings
were evaluated on $1{,}000$ weak-contour carriers.
Table~\ref{tab:sweep} reports each operation's largest available sample;
its $n$ column gives that sample size, not the screening batch.

\begin{table}[!htbp]
\caption{All $27$ tested operations on contours at $s=1.0$. The $n$ column gives
the sample size for the reported result; rows with different sample sizes
should not be compared at subpercentage-point precision. Operations are
grouped into four families. A dash denotes no parameter sweep.}
\label{tab:sweep}
\begin{center}
\setlength{\tabcolsep}{2.5pt}
\begin{tabular}{llrr}
\toprule
Family and grid & Setting & Contour $s{=}1.0$ $\uparrow$ & $n$ \\
\midrule
\multicolumn{4}{l}{\textbf{low-pass} (16 operations)} \\
Gaussian blur, $\sigma \in \{4,8,16,24,32,48\}$ & $\sigma=16$ & \best{15.00} & $1{,}000$ \\
                                       & $\sigma=8$  & 11.00 & $1{,}000$ \\
                                       & $\sigma=24$ & 5.30 & $1{,}000$ \\
                                       & $\sigma=4$  & 2.33 & 300 \\
                                       & $\sigma=32$ & 2.10 & $1{,}000$ \\
                                       & $\sigma=48$ & 2.10 & $1{,}000$ \\
downscale, $f \in \{4,6,8,16,32,64\}$  & $f=32$ \ (16\,px) & 6.00 & 300 \\
                                       & $f=16$ \ (32\,px) & 5.67 & 300 \\
                                       & $f=64$ \ (8\,px)  & 1.70 & $1{,}000$ \\
                                       & $f=4$ \ (128\,px) & 1.40 & $1{,}000$ \\
                                       & $f=6$ \ (85\,px)  & 1.10 & $1{,}000$ \\
                                       & $f=8$ \ (64\,px)  & 1.00 & 300 \\
FFT low-pass, $r \in \{12,25,50,100\}$ & $r=25$  & 8.67 & 300 \\
                                       & $r=50$  & 4.00 & 300 \\
                                       & $r=12$  & 1.50 & $1{,}000$ \\
                                       & $r=100$ & 1.33 & 300 \\
\midrule
\multicolumn{4}{l}{\textbf{combination} (2 operations)} \\
blur then equalize, $\sigma \in \{8\}$ \citep{qu2025hate} & $\sigma=8$ & 10.10 & $1{,}000$ \\
downscale $1/16$ then Otsu             & $f=16$ \ (32\,px) & 10.33 & 300 \\
\midrule
\multicolumn{4}{l}{\textbf{pointwise} (6 operations)} \\
Otsu threshold \citep{otsu1979threshold}        & ---       & 2.00 & $1{,}000$ \\
adaptive threshold, block $\in \{51\}$ & Block size 51 & 2.00 & 300 \\
histogram equalization                 & ---     & 1.67 & 300 \\
percentile stretch, $p \in \{2\}$      & $p=2$ & 1.67 & 300 \\
grayscale                              & ---       & 1.33 & 300 \\
CLAHE, clip $\in \{3\}$                & Clip limit 3  & 1.33 & 300 \\
\midrule
\multicolumn{4}{l}{\textbf{edge} (3 operations)} \\
Canny edges                            & ---      & 1.33 & 300 \\
unsharp mask, $\sigma \in \{8\}$       & $\sigma=8$ & 1.33 & 300 \\
Sobel magnitude                        & ---      & 1.00 & $1{,}000$ \\
\bottomrule
\end{tabular}
\end{center}
\end{table}

Within the tested grids, peak scores occur away from the endpoints
(Figure~\ref{fig:recovery_analysis}). Gaussian blur reaches $15.00\%$
at $\sigma=16$ ($95\%$ item-bootstrap CI, $[9.80,20.60]$), compared with
$11.00\%$ at $\sigma=8$ ($[7.40,15.00]$); these intervals do not resolve
which setting is better. For downsampling, the $0.33$-point difference
between $16$ and $32$ pixels does not establish an advantage outside
SemVink's stated $32$--$128$-pixel range. The best tested single operation
remains below \method{} on weak contours.

These results cover the tested grids rather than an upper bound on
image transformations. Both the fixed-operation settings and the learned
configuration were selected using downstream recognition accuracy, so
configuration selection can affect both sides of the comparison. The
intervals quantify item variation conditional on the selected settings
and do not include this selection uncertainty. AVR's view parameters
were fixed before recognition evaluation; its composite seven-view
protocol is separate from the $27$ single-operation sweep.

\section{Reader and Generator Generalization}
\label{app:generalization}

This appendix provides the reader and generator comparisons supporting Section~\ref{sec:readers}, together with clean-field recognition references. The generator tables separate five QR-based hidden-content configurations from an auxiliary edge-control reference. The tables report single-answer accuracy, with complete AVR seven-answer coverage diagnostics retained in the accompanying Source Data.

\subsection{Reader Generalization}
\label{app:readers}

At $s=1.0$, \method{} gains approximately $14$ points over SMSP on words
with three readers; InternVL3-8B instead gives a $-5.33$-point difference
whose interval contains zero. Across strengths, the differences from SMSP
are $-0.89$ points for Qwen2.5-VL-7B, $-9.44$ for InternVL3-8B,
$+11.33$ for Qwen2.5-VL-32B and $+11.00$ for
LLaVA-OneVision-7B; the paired intervals for the latter three differences
exclude zero. Thus even the aggregate word comparison depends on
the reader.

Across strengths, \method{}'s word accuracy exceeds AVR's selected-answer accuracy and its separate seven-answer coverage for all four readers. Individual conditions can reverse the coverage comparison: with InternVL3-8B at $s=2.0$, AVR any-view coverage reaches $82.67\%$, compared with $80.33\%$ for \method{}.

Tables~\ref{tab:readers_full_pattern} and~\ref{tab:readers_full_word} report SMSP, AVR and \method{} on the same $1{,}800$ carriers for each reader, with $300$ per domain--strength cell. The word domain combines real words and non-words. All four readers use the same fixed recovery checkpoint trained with seed $42$; these results are separate from the three-seed main comparison.

For Qwen2.5-VL-7B, InternVL3-8B, Qwen2.5-VL-32B and LLaVA-OneVision-7B, direct weak-contour recognition is $1.67\%$, $1.67\%$, $2.33\%$ and $0.67\%$, respectively. The corresponding \method{} gains over SMSP are $28.00$, $15.00$, $32.00$ and $39.33$ percentage points; all paired intervals exclude zero. Clean-field references appear in Table~\ref{tab:clean_readers}.

\begin{table}[!htbp]
\centering
\small
\setlength{\tabcolsep}{4pt}
\caption{Contour recognition accuracy (\%) across readers, with $300$ carriers per strength and the same recovery checkpoint (training seed $42$). AVR uses plurality voting.}
\label{tab:readers_full_pattern}
\begin{tabular}{llrrr}
\toprule
\noalign{\vskip 4.30871pt}
Reader & $s$ & SMSP $\uparrow$ & AVR $\uparrow$ & \textbf{ControlTrace (Ours)} $\uparrow$ \\
\noalign{\vskip 4.30871pt}
\midrule
Qwen2.5-VL-7B & 1.0 & 15.33 & 3.67 & 43.33 \\
 & 1.5 & 41.00 & 18.33 & 65.00 \\
 & 2.0 & 48.33 & 36.67 & 69.67 \\
\addlinespace[3pt]
InternVL3-8B & 1.0 & 12.67 & 3.00 & 27.67 \\
 & 1.5 & 30.67 & 13.00 & 48.33 \\
 & 2.0 & 38.33 & 22.33 & 49.00 \\
\addlinespace[3pt]
Qwen2.5-VL-32B & 1.0 & 7.67 & 2.00 & 39.67 \\
 & 1.5 & 26.00 & 16.33 & 58.67 \\
 & 2.0 & 41.33 & 29.67 & 65.33 \\
\addlinespace[3pt]
LLaVA-OneVision-7B & 1.0 & 7.00 & 5.33 & 46.33 \\
 & 1.5 & 21.00 & 31.33 & 73.67 \\
 & 2.0 & 38.00 & 56.33 & 81.33 \\
\bottomrule
\end{tabular}
\end{table}

\begin{table}[!htbp]
\centering
\small
\setlength{\tabcolsep}{4pt}
\caption{Word recognition accuracy (\%) across readers, with $300$ carriers per strength and the same recovery checkpoint (training seed $42$). AVR uses plurality voting.}
\label{tab:readers_full_word}
\begin{tabular}{llrrr}
\toprule
\noalign{\vskip 4.30871pt}
Reader & $s$ & SMSP $\uparrow$ & AVR $\uparrow$ & \textbf{ControlTrace (Ours)} $\uparrow$ \\
\noalign{\vskip 4.30871pt}
\midrule
Qwen2.5-VL-7B & 1.0 & 14.33 & 1.00 & 28.33 \\
 & 1.5 & 79.67 & 16.33 & 76.67 \\
 & 2.0 & 93.67 & 49.00 & 80.00 \\
\addlinespace[3pt]
InternVL3-8B & 1.0 & 27.00 & 1.33 & 21.67 \\
 & 1.5 & 85.67 & 18.67 & 76.67 \\
 & 2.0 & 94.33 & 60.00 & 80.33 \\
\addlinespace[3pt]
Qwen2.5-VL-32B & 1.0 & 8.67 & 0.33 & 22.67 \\
 & 1.5 & 64.00 & 16.67 & 81.33 \\
 & 2.0 & 85.00 & 48.67 & 87.67 \\
\addlinespace[3pt]
LLaVA-OneVision-7B & 1.0 & 14.00 & 1.33 & 28.67 \\
 & 1.5 & 60.67 & 20.67 & 76.00 \\
 & 2.0 & 77.33 & 55.67 & 80.33 \\
\bottomrule
\end{tabular}
\end{table}

\subsection{Clean-Field Recognition}
\label{app:clean_readers}

Table~\ref{tab:clean_readers} reports direct recognition of clean control
fields associated with the $s=1.0$ subset of the reader evaluation.
Word accuracy exceeds contour accuracy by $2$--$38$ points across the
four readers. These scores provide reader references; the paired anagram
comparison in Section~\ref{sec:lexical} examines lexical content more directly.

\begin{table}[!htbp]
\centering
\small
\caption{Direct recognition of clean control fields (\%) associated with the
$s=1.0$ subset of the reader evaluation.}
\label{tab:clean_readers}
\begin{tabular}{lrr}
\toprule
Reader & Contours $\uparrow$ & Words $\uparrow$ \\
\midrule
Qwen2.5-VL-7B & 78.0 & 80.0 \\
InternVL3-8B & 56.0 & 94.0 \\
Qwen2.5-VL-32B & 78.0 & 92.0 \\
LLaVA-OneVision-7B & 90.0 & 94.0 \\
\bottomrule
\end{tabular}
\end{table}

\subsection{Generator Transfer}
\label{app:transfer}

We evaluate transfer without adaptation on five QR-based generator configurations, including SD1.5 with QR Code Monster, which also supplies the training pairs. Canny ControlNet uses a different control objective and remains an auxiliary reference. Table~\ref{tab:transfer_full_pattern} reports contour recognition by strength and across all $900$ carriers per configuration; the SDXL backbone follows \citet{podell2024sdxl}.

Among the five QR-based configurations, \method{} has higher aggregate contour accuracy than SMSP in three configurations and than AVR vote in four (Table~\ref{tab:transfer_full_pattern}). QR Code Monster with SD1.5 yields $63.78\%$, compared with $37.56\%$ for SMSP and $22.56\%$ for AVR vote. The seven-answer diagnostic reaches $47.33\%$ in this configuration; it exceeds \method{} at $s=2.0$ on SDXL ($65.00\%$ versus $54.33\%$) and QR Pattern ($69.67\%$ versus $60.00\%$).

The QR Code configurations show a larger transfer failure.
On SD1.5, \method{} reaches $4.78\%$ across strengths, below SMSP's
$22.56\%$, AVR plurality's $11.56\%$ and AVR any-view coverage of
$27.89\%$. On SD2.1, \method{} reaches $27.56\%$, below SMSP's
$30.11\%$ and AVR coverage of $37.44\%$. Increasing conditioning strength
does not resolve the SD1.5 failure: at $s=2.0$, \method{} reaches $6.33\%$
while SMSP reaches $38.00\%$ and AVR coverage reaches $47.33\%$.

Word recognition also limits transfer. Across strengths, AVR any-view
coverage exceeds \method{} on QR Pattern and both
QR Code configurations; its plurality readout exceeds \method{}
on both QR Code configurations. The single-answer comparisons appear in Tables~\ref{tab:transfer_full_pattern} and~\ref{tab:transfer_full_word}.
These results support sensitivity to the encoding process, rather than
a general transfer advantage of learned recovery.

In the auxiliary Canny configuration, \method{} reaches $16.00\%$ contour accuracy, compared with $34.33\%$ for direct reading, $38.89\%$ for SMSP and $35.33\%$ for AVR vote; AVR any-view coverage is $54.67\%$. Word accuracy is $5.44\%$ for \method{} and $45.67\%$ for AVR vote, with $56.56\%$ any-view coverage. These results describe transfer to edge-based control and are separate from the QR-based hidden-content comparison.

Tables~\ref{tab:transfer_full_pattern} and~\ref{tab:transfer_full_word} give both domains at each strength, with $300$ carriers per cell. Each method reads the same carriers, and \method{} uses one recovery checkpoint without adaptation. Direct denotes recognition from the unmodified carrier.

For the five QR configurations in table order, the aggregate \method{}--SMSP differences are $+26.2$, $+5.2$, $+12.6$, $-17.8$ and $-2.6$ percentage points. At $s=1.0$, they are $+38.3$, $+15.3$, $+21.0$, $-2.3$ and $+1.7$ points. Paired intervals exclude zero for the first, third and fourth aggregate differences and the first three weak-condition differences. The auxiliary Canny differences are $-22.9$ points overall and $-14.3$ points at $s=1.0$.

\begin{table}[!htbp]
\centering
\small
\setlength{\tabcolsep}{3.5pt}
\caption{Contour recognition accuracy (\%) across generators, with $300$ carriers per strength and one recovery checkpoint. The All rows aggregate the $900$ carriers across strengths. SD1.5 and QR Code Monster also generate the training pairs; Canny is an auxiliary edge-control reference. AVR uses plurality voting.}
\label{tab:transfer_full_pattern}
\begin{tabular}{llrrrrr}
\toprule
\noalign{\vskip 4.30872pt}
Configuration & $s$ & Direct $\uparrow$ & SemVink $\uparrow$ & SMSP $\uparrow$ & AVR $\uparrow$ & \textbf{ControlTrace (Ours)} $\uparrow$ \\
\noalign{\vskip 4.30872pt}
\midrule
SD1.5 / QR Code Monster & 1.0 & 1.33 & 2.33 & 12.67 & 4.00 & 51.00 \\
 & 1.5 & 10.33 & 17.67 & 44.33 & 21.00 & 68.33 \\
 & 2.0 & 19.00 & 33.33 & 55.67 & 42.67 & 72.00 \\
 & All & 10.22 & 17.78 & 37.56 & 22.56 & 63.78 \\
\addlinespace[3pt]
SDXL / QR Code Monster & 1.0 & 2.00 & 3.00 & 12.33 & 2.00 & 27.67 \\
 & 1.5 & 6.33 & 16.33 & 39.00 & 14.33 & 48.00 \\
 & 2.0 & 33.00 & 46.33 & 63.00 & 41.67 & 54.33 \\
 & All & 13.78 & 21.89 & 38.11 & 19.33 & 43.33 \\
\addlinespace[3pt]
SD1.5 / QR Pattern & 1.0 & 1.00 & 1.67 & 3.33 & 0.33 & 24.33 \\
 & 1.5 & 9.00 & 9.00 & 20.67 & 12.33 & 31.00 \\
 & 2.0 & 31.33 & 39.33 & 53.67 & 48.67 & 60.00 \\
 & All & 13.78 & 16.67 & 25.89 & 20.44 & 38.44 \\
\addlinespace[3pt]
SD1.5 / QR Code & 1.0 & 3.00 & 1.33 & 5.00 & 2.00 & 2.67 \\
 & 1.5 & 2.00 & 11.67 & 24.67 & 8.67 & 5.33 \\
 & 2.0 & 10.67 & 23.67 & 38.00 & 24.00 & 6.33 \\
 & All & 5.22 & 12.22 & 22.56 & 11.56 & 4.78 \\
\addlinespace[3pt]
SD2.1 / QR Code & 1.0 & 1.67 & 1.33 & 2.00 & 1.67 & 3.67 \\
 & 1.5 & 8.67 & 16.33 & 32.67 & 17.33 & 30.00 \\
 & 2.0 & 20.33 & 38.00 & 55.67 & 38.00 & 49.00 \\
 & All & 10.22 & 18.56 & 30.11 & 19.00 & 27.56 \\
\midrule
\multicolumn{7}{l}{\textbf{Auxiliary edge-control reference}} \\
SD1.5 / Canny & 1.0 & 19.00 & 8.00 & 20.00 & 17.00 & 5.67 \\
 & 1.5 & 38.00 & 24.33 & 43.33 & 38.33 & 16.67 \\
 & 2.0 & 46.00 & 37.00 & 53.33 & 50.67 & 25.67 \\
 & All & 34.33 & 23.11 & 38.89 & 35.33 & 16.00 \\
\bottomrule
\end{tabular}
\end{table}

\begin{table}[!htbp]
\centering
\small
\setlength{\tabcolsep}{3.5pt}
\caption{Word recognition accuracy (\%) across generators, with $300$ carriers per strength and one recovery checkpoint. Canny is an auxiliary edge-control reference. AVR uses plurality voting.}
\label{tab:transfer_full_word}
\begin{tabular}{llrrrrr}
\toprule
\noalign{\vskip 4.30872pt}
Configuration & $s$ & Direct $\uparrow$ & SemVink $\uparrow$ & SMSP $\uparrow$ & AVR $\uparrow$ & \textbf{ControlTrace (Ours)} $\uparrow$ \\
\noalign{\vskip 4.30872pt}
\midrule
SD1.5 / QR Code Monster & 1.0 & 0.00 & 14.67 & 31.00 & 1.33 & 40.67 \\
 & 1.5 & 1.67 & 69.33 & 78.33 & 20.67 & 73.33 \\
 & 2.0 & 14.33 & 85.00 & 92.00 & 58.67 & 81.33 \\
\addlinespace[3pt]
SDXL / QR Code Monster & 1.0 & 0.00 & 18.33 & 29.67 & 1.67 & 25.33 \\
 & 1.5 & 6.33 & 72.33 & 82.00 & 28.00 & 65.33 \\
 & 2.0 & 40.67 & 80.00 & 96.00 & 72.00 & 72.67 \\
\addlinespace[3pt]
SD1.5 / QR Pattern & 1.0 & 0.00 & 15.00 & 24.33 & 1.67 & 26.67 \\
 & 1.5 & 0.33 & 20.33 & 23.33 & 6.67 & 13.33 \\
 & 2.0 & 20.33 & 67.00 & 78.33 & 57.33 & 46.00 \\
\addlinespace[3pt]
SD1.5 / QR Code & 1.0 & 1.00 & 2.67 & 4.67 & 2.33 & 0.00 \\
 & 1.5 & 6.67 & 39.33 & 48.33 & 18.67 & 2.33 \\
 & 2.0 & 22.33 & 63.00 & 77.00 & 42.67 & 5.00 \\
\addlinespace[3pt]
SD2.1 / QR Code & 1.0 & 0.33 & 0.67 & 2.67 & 0.67 & 0.67 \\
 & 1.5 & 10.33 & 43.67 & 56.00 & 28.67 & 35.00 \\
 & 2.0 & 32.00 & 76.67 & 90.00 & 70.33 & 61.33 \\
\midrule
\multicolumn{7}{l}{\textbf{Auxiliary edge-control reference}} \\
SD1.5 / Canny & 1.0 & 6.00 & 2.33 & 6.00 & 5.67 & 1.00 \\
 & 1.5 & 52.67 & 23.67 & 58.00 & 54.67 & 6.33 \\
 & 2.0 & 72.67 & 42.33 & 78.00 & 76.67 & 9.00 \\
\bottomrule
\end{tabular}
\end{table}

\section{Robustness}
\label{app:robustness}

This appendix expands the ControlTrace perturbation results in Section~\ref{sec:readers} and Figure~\ref{fig:robustness_unified}. The tables report single-answer accuracy at all tested strengths in both domains, with methods evaluated on the same perturbed carriers. The corresponding AVR seven-answer coverage diagnostics are retained in the accompanying Source Data.

\subsection{Common Image Perturbations}
\label{app:robust}

To test common image changes during distribution, we apply JPEG compression, Gaussian noise or downsampling followed by restoration to the original dimensions. Gaussian noise has a standard deviation of $10$ in 8-bit intensity units ($10/255$ on $[0,1]$). Each method receives the same perturbed carrier; AVR constructs its seven views afterwards. Each condition contains $1{,}800$ carriers, with $300$ per domain and strength. Tables~\ref{tab:robustness_full_pattern} and~\ref{tab:robustness_full_word} report all strengths using the same single-checkpoint Qwen2.5-VL-7B subset as the reader comparison.

At weak conditioning, \method{} reaches $41.3$--$44.0\%$ contour accuracy after these perturbations. Across the two domains, its largest absolute change from the unperturbed score is $2.33$ points; the paired intervals include zero. AVR any-view coverage ranges from $16.7\%$ to $19.0\%$ on contours and from $3.7\%$ to $5.3\%$ on words.

Across strengths, \method{} remains within one point of its unperturbed
score in both domains. AVR any-view coverage is also similar across these
conditions: $43.2$--$44.6\%$ for contours and $42.0$--$43.3\%$ for words,
compared with $43.3\%$ and $41.7\%$ before perturbation. Its single-answer
readout changes more under downsampling, with word accuracy rising from
$22.1\%$ to $27.9\%$. At $s=2.0$ after downsampling, word coverage
reaches $80.3\%$, slightly above \method{}'s $79.3\%$.

Direct weak-contour accuracy is $1.67\%$ without perturbation and $1.33\%$, $2.00\%$, $2.00\%$ and $1.00\%$ after JPEG ($q=75$), JPEG ($q=50$), noise and downsampling, respectively. Direct weak-word accuracy is $0.33\%$ in all five conditions. Across strengths, downsampling raises direct word recognition from $2.56\%$ to $6.44\%$, showing that the operation also affects recognition without a recovery step.

\begin{table}[!htbp]
\centering
\small
\setlength{\tabcolsep}{4pt}
\caption{Contour recognition accuracy (\%) under common image perturbations, with $300$ carriers per strength and one recovery checkpoint. AVR uses plurality voting.}
\label{tab:robustness_full_pattern}
\begin{tabular}{llrrr}
\toprule
\noalign{\vskip 4.30871pt}
Configuration & $s$ & SMSP $\uparrow$ & AVR $\uparrow$ & \textbf{ControlTrace (Ours)} $\uparrow$ \\
\noalign{\vskip 4.30871pt}
\midrule
Unperturbed & 1.0 & 15.33 & 3.67 & 43.33 \\
 & 1.5 & 41.00 & 18.33 & 65.00 \\
 & 2.0 & 48.33 & 36.67 & 69.67 \\
\addlinespace[3pt]
JPEG ($q=75$) & 1.0 & 16.00 & 5.00 & 44.00 \\
 & 1.5 & 39.67 & 18.67 & 66.33 \\
 & 2.0 & 49.00 & 36.67 & 69.33 \\
\addlinespace[3pt]
JPEG ($q=50$) & 1.0 & 14.33 & 2.67 & 42.33 \\
 & 1.5 & 40.33 & 19.33 & 66.33 \\
 & 2.0 & 50.33 & 37.67 & 70.67 \\
\addlinespace[3pt]
Gaussian noise ($\sigma=10$) & 1.0 & 16.33 & 3.67 & 41.67 \\
 & 1.5 & 39.00 & 20.00 & 64.67 \\
 & 2.0 & 50.33 & 38.67 & 71.00 \\
\addlinespace[3pt]
Downsample $4\times$ & 1.0 & 15.67 & 3.67 & 41.33 \\
 & 1.5 & 41.00 & 20.33 & 66.33 \\
 & 2.0 & 51.00 & 40.67 & 70.67 \\
\bottomrule
\end{tabular}
\end{table}

\begin{table}[!htbp]
\centering
\small
\setlength{\tabcolsep}{4pt}
\caption{Word recognition accuracy (\%) under common image perturbations, with $300$ carriers per strength and one recovery checkpoint. AVR uses plurality voting.}
\label{tab:robustness_full_word}
\begin{tabular}{llrrr}
\toprule
\noalign{\vskip 4.30871pt}
Configuration & $s$ & SMSP $\uparrow$ & AVR $\uparrow$ & \textbf{ControlTrace (Ours)} $\uparrow$ \\
\noalign{\vskip 4.30871pt}
\midrule
Unperturbed & 1.0 & 14.33 & 1.00 & 28.33 \\
 & 1.5 & 79.67 & 16.33 & 76.67 \\
 & 2.0 & 93.67 & 49.00 & 80.00 \\
\addlinespace[3pt]
JPEG ($q=75$) & 1.0 & 15.33 & 0.33 & 29.67 \\
 & 1.5 & 78.67 & 15.33 & 77.00 \\
 & 2.0 & 94.00 & 48.67 & 79.67 \\
\addlinespace[3pt]
JPEG ($q=50$) & 1.0 & 13.67 & 0.67 & 28.67 \\
 & 1.5 & 78.33 & 18.00 & 76.00 \\
 & 2.0 & 94.00 & 47.67 & 79.33 \\
\addlinespace[3pt]
Gaussian noise ($\sigma=10$) & 1.0 & 16.33 & 0.67 & 28.67 \\
 & 1.5 & 80.00 & 17.67 & 77.67 \\
 & 2.0 & 93.67 & 49.33 & 80.00 \\
\addlinespace[3pt]
Downsample $4\times$ & 1.0 & 15.67 & 1.00 & 26.00 \\
 & 1.5 & 82.67 & 22.33 & 78.00 \\
 & 2.0 & 92.67 & 60.33 & 79.33 \\
\bottomrule
\end{tabular}
\end{table}

\subsection{Adversarial Evasion}
\label{app:adversarial}

\begin{table}[!htbp]
\centering
\small
\setlength{\tabcolsep}{4pt}
\caption{Contour recognition accuracy (\%) under perturbations and attacks, with $300$ carriers per strength and one recovery checkpoint. All methods read the same perturbed carriers. AVR uses plurality voting; Blur denotes the auxiliary Gaussian-blur reference ($\sigma=16$).}
\label{tab:adversarial_full_pattern}
\begin{tabular}{llrrrr}
\toprule
\noalign{\vskip 4.30871pt}
Configuration & $s$ & Blur $\uparrow$ & SMSP $\uparrow$ & AVR $\uparrow$ & \textbf{ControlTrace (Ours)} $\uparrow$ \\
\noalign{\vskip 4.30871pt}
\midrule
Band noise ($0.01$--$0.08$) & 1.0 & 13.33 & 15.67 & 3.33 & 42.00 \\
 & 1.5 & 41.00 & 41.67 & 20.33 & 66.00 \\
 & 2.0 & 49.33 & 51.00 & 35.33 & 70.00 \\
\addlinespace[3pt]
Band noise ($0.08$--$0.25$) & 1.0 & 13.33 & 17.00 & 5.00 & 43.67 \\
 & 1.5 & 39.67 & 40.00 & 20.33 & 66.67 \\
 & 2.0 & 48.67 & 51.00 & 40.00 & 70.00 \\
\addlinespace[3pt]
Band noise ($0.25$--$0.50$) & 1.0 & 14.00 & 15.67 & 3.00 & 42.67 \\
 & 1.5 & 40.00 & 39.67 & 20.67 & 65.67 \\
 & 2.0 & 49.67 & 51.33 & 36.67 & 69.67 \\
\addlinespace[3pt]
PGD ($4/255$) & 1.0 & 12.33 & 12.67 & 3.67 & 20.33 \\
 & 1.5 & 38.67 & 39.00 & 18.00 & 50.33 \\
 & 2.0 & 48.67 & 49.00 & 39.00 & 57.00 \\
\addlinespace[3pt]
PGD ($8/255$) & 1.0 & 10.33 & 11.00 & 4.00 & 9.00 \\
 & 1.5 & 36.00 & 37.00 & 16.33 & 35.00 \\
 & 2.0 & 47.00 & 49.67 & 38.67 & 51.33 \\
\addlinespace[3pt]
PGD ($8/255$) + JPEG & 1.0 & 9.33 & 10.67 & 2.00 & 9.67 \\
 & 1.5 & 36.00 & 37.33 & 17.33 & 34.67 \\
 & 2.0 & 47.00 & 48.33 & 38.33 & 51.00 \\
\bottomrule
\end{tabular}
\end{table}

\begin{table}[!htbp]
\centering
\small
\setlength{\tabcolsep}{4pt}
\caption{Word recognition accuracy (\%) under perturbations and attacks, with $300$ carriers per strength and one recovery checkpoint. All methods read the same perturbed carriers. AVR uses plurality voting; Blur denotes the auxiliary Gaussian-blur reference ($\sigma=16$).}
\label{tab:adversarial_full_word}
\begin{tabular}{llrrrr}
\toprule
\noalign{\vskip 4.30871pt}
Configuration & $s$ & Blur $\uparrow$ & SMSP $\uparrow$ & AVR $\uparrow$ & \textbf{ControlTrace (Ours)} $\uparrow$ \\
\noalign{\vskip 4.30871pt}
\midrule
Band noise ($0.01$--$0.08$) & 1.0 & 6.33 & 16.67 & 0.33 & 26.00 \\
 & 1.5 & 46.00 & 77.33 & 19.33 & 78.67 \\
 & 2.0 & 62.67 & 92.33 & 51.67 & 81.33 \\
\addlinespace[3pt]
Band noise ($0.08$--$0.25$) & 1.0 & 6.33 & 15.33 & 1.00 & 29.00 \\
 & 1.5 & 46.33 & 80.00 & 19.33 & 77.00 \\
 & 2.0 & 64.67 & 93.67 & 57.33 & 79.33 \\
\addlinespace[3pt]
Band noise ($0.25$--$0.50$) & 1.0 & 6.00 & 16.00 & 1.00 & 29.33 \\
 & 1.5 & 48.00 & 81.00 & 17.00 & 77.00 \\
 & 2.0 & 65.33 & 93.00 & 52.33 & 79.67 \\
\addlinespace[3pt]
PGD ($4/255$) & 1.0 & 6.00 & 12.67 & 0.00 & 6.33 \\
 & 1.5 & 45.67 & 75.67 & 17.33 & 46.67 \\
 & 2.0 & 62.67 & 93.33 & 50.00 & 66.33 \\
\addlinespace[3pt]
PGD ($8/255$) & 1.0 & 4.33 & 10.33 & 1.00 & 1.00 \\
 & 1.5 & 44.33 & 74.33 & 14.67 & 15.33 \\
 & 2.0 & 63.00 & 92.33 & 50.33 & 42.33 \\
\addlinespace[3pt]
PGD ($8/255$) + JPEG & 1.0 & 4.67 & 9.67 & 0.33 & 1.00 \\
 & 1.5 & 43.67 & 73.00 & 14.67 & 17.00 \\
 & 2.0 & 63.00 & 92.33 & 54.00 & 44.33 \\
\bottomrule
\end{tabular}
\end{table}

Band-limited noise uses frequency intervals $[0.01,0.08)$, $[0.08,0.25)$ and $[0.25,0.50)$ cycles per pixel, with a per-channel standard deviation of $12/255$. For the white-box attack, let $I_{\mathrm{r}}$ denote the carrier on the $256\times256$ recovery-input grid, and let $f$ denote the grayscale recovery mapping. PGD maximizes $\|f(I_{\mathrm{r}}+\delta)-c\|_1$ subject to $\|\delta\|_\infty\leq\varepsilon$, for $\varepsilon\in\{4/255,8/255\}$. Both budgets use $20$ steps of size $1/255$, uniform random initialization within the perturbation budget, and projection followed by clipping to $[0,1]$ at each step. The perturbed input is upsampled to $512\times512$ before saving; the stated budget applies on the recovery-input grid. All compared methods receive these same saved carriers, so SMSP and AVR measure transfer of the recovery-targeted attack.

Tables~\ref{tab:adversarial_full_pattern} and~\ref{tab:adversarial_full_word} report all tested strengths, with $300$ carriers per domain--strength cell and one recovery checkpoint. Gaussian blur ($\sigma=16$) provides an auxiliary VLM readability diagnostic, with unperturbed weak-contour accuracy of $13.67\%$. This diagnostic does not measure human readability after attack. Unperturbed SMSP, AVR and \method{} scores appear in the Qwen2.5-VL-7B rows of Tables~\ref{tab:readers_full_pattern} and~\ref{tab:readers_full_word}.

The low-, mid- and high-band perturbations change \method{}'s weak-contour accuracy by $-1.33$, $+0.33$ and $-0.67$ percentage points, respectively. PGD reduces it by $23.00$ points at $4/255$ and $34.33$ points at $8/255$; the reduction after JPEG recompression is $33.67$ points. The paired intervals exclude zero for all three attack-related reductions. At $8/255$, AVR any-view coverage is $15.33\%$, or $14.67\%$ after JPEG, while its plurality accuracy is $4.00\%$ and $2.00\%$, respectively.

Across strengths, the $8/255$ attack reduces \method{} to $31.78\%$
contour accuracy, compared with AVR coverage of $42.33\%$.
On words, \method{} reaches $19.56\%$, AVR plurality $22.00\%$ and
coverage $39.00\%$; SMSP remains at $59.00\%$. After JPEG recompression,
the corresponding word scores are $20.78\%$, $23.00\%$, $39.22\%$ and
$58.33\%$. Tables~\ref{tab:adversarial_full_pattern} and~\ref{tab:adversarial_full_word} give the single-answer comparisons.

\section{Specificity and Qualitative Results}
\label{app:specificity_qualitative}

The negative controls and recovered-field statistics supplement Section~\ref{sec:specificity}. The qualitative examples expand the recognition results in Section~\ref{sec:main} and Figure~\ref{fig:teaser}.

\subsection{Negative Controls}
\label{app:negatives}

\begin{table}[!htbp]
\centering
\small
\setlength{\tabcolsep}{3.5pt}
\caption{Content-reporting rates (\%) under the shared none-capable prompt. Positive-image reports need not identify the correct target; reports on negative images are false positives. ControlTrace uses one checkpoint, with $95\%$ intervals in brackets. All four methods produced valid final answers, and both rejection parsers agreed. AVR uses its plurality-selected answer.}
\label{tab:specificity}
\begin{tabular}{lrrrrr}
\toprule
\noalign{\vskip 4.28419pt}
Image set & $n$ & SemVink & SMSP & AVR & \textbf{ControlTrace (Ours)} \\
\noalign{\vskip 4.28419pt}
\midrule
\multicolumn{6}{l}{\textbf{Positive images: content reporting}} \\
Hidden contours & $1{,}000$ & 3.50 & 9.60 & 0.10 & 61.5\ [53.1, 69.8] \\
\midrule
\multicolumn{6}{l}{\textbf{Negative images: false-positive rate $\downarrow$}} \\
Matched, control off & $1{,}000$ & 2.30 & 0.40 & 0.00 & 3.8\ [2.4, 5.5] \\
Dense-texture scenes & $1{,}000$ & 1.30 & 2.00 & 0.40 & 5.2\ [3.1, 7.6] \\
COCO photographs & $10{,}000$ & 10.54 & 9.04 & 1.45 & 5.3\ [4.8, 5.7] \\
Non-semantic conditioning & $960$ & 2.81 & 1.15 & 0.00 & 2.9\ [1.8, 4.3] \\
\bottomrule
\end{tabular}
\end{table}

\paragraph{Photographs.} We use all $5{,}000$ images of the COCO 2017 validation set and $5{,}000$ images sampled from its test set with seed $20260912$. Each image is center-cropped, resized to $512\times512$ with Lanczos resampling and saved as PNG to match the carrier resolution and format.

\paragraph{Matched counterfactuals.} For each of the $1{,}000$ contour carriers at $s=1.0$, we generate a negative image with the same base model, scheduler, step count, guidance scale, scene prompt, negative prompt and seed, using a blank control field and zero conditioning strength. Seeds are unique across the set, and each negative is paired with one positive.

\begin{figure}[!htbp]
\begin{center}
\includegraphics[width=\linewidth,height=0.68\textheight,keepaspectratio]{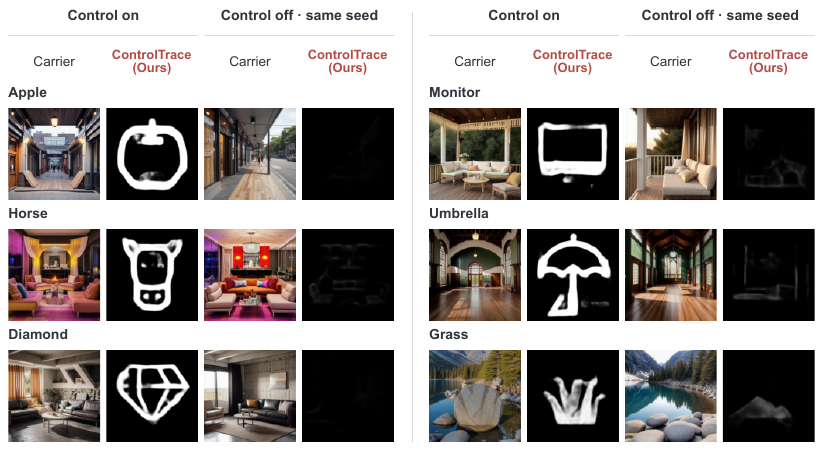}
\end{center}
\caption{Configuration-matched positive and negative carriers, with their recovered
fields. Six targets were sampled with seed $20260913$, one scene per target,
from $1{,}000$ matched pairs. Each pair shares the seed, prompt, scheduler,
steps and guidance, with the control branch on or off. Removing control
residuals changes the generation trajectory, so the match is by
configuration rather than by pixels.}
\label{fig:matched}
\end{figure}

\paragraph{Non-semantic conditioning.} To test active conditioning without a text or object-contour target, we generated $960$ carriers at $s=1.0$ from $48$ fields. The fields cover eight families: QR-like block lattices with finder squares, checkerboards, concentric rings, stripes, maze hatching, thresholded low-frequency blobs, halftone dot grids, and spirals. These geometric patterns test whether imposed structure alone elicits false reports of meaningful hidden content. Six fields per family were generated from a stated seed. Each field uses the twenty scenes and prompts of one contour item, matching the positive scene distribution. The sampling unit is the field because responses across its scenes are not independent.

\paragraph{Dense-texture scenes.} Twenty scene prompts chosen for repetitive,
high-contrast, contour-rich content (weathered brick, overlapping foliage,
cobblestone, patterned tile, turbulent cloud, forest canopy, rocky shore, market
crowd, packed bookshelf, circuit board, bare branches, coarse weave, cracked
paint, gravel, chain-link fence, stacked firewood, coral, trampled snow, torn
posters, rusted metal), fifty seeds each, generated the same way with control
off.

\subsection{Recovered-field Statistics}
\label{app:fieldstats}

We computed gradient energy from the saved grayscale outputs of the seed-$42$ checkpoint, using $1{,}000$ positive contour carriers at $s=1.0$. For an output $f$ clipped to $[0,1]$, the score is the mean magnitude of two-pixel central differences:

\begin{equation}
\begin{aligned}
\mathcal{E}_{\mathrm{grad}}(f)&=\frac{1}{HW}\sum_{i,j}\sqrt{d_x(i,j)^2+d_y(i,j)^2},\\
d_x(i,j)&=f_{i,j+1}-f_{i,j-1},\qquad
d_y(i,j)=f_{i+1,j}-f_{i-1,j}.
\end{aligned}
\end{equation}

\begin{figure}[!htbp]
\centering
\includegraphics[width=\linewidth]{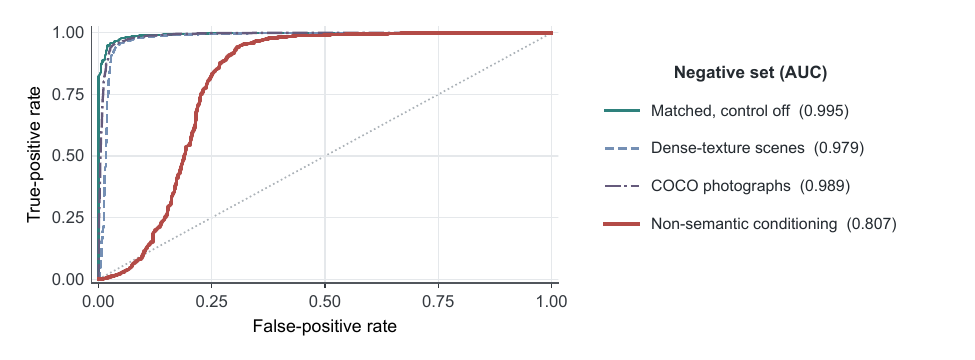}
\caption{Empirical ROC curves for recovered-field gradient energy. All curves share $1{,}000$ positive contours. Negative sets comprise matched control-off images ($n=1{,}000$, AUC $0.9947$), dense-texture scenes ($n=1{,}000$, $0.9792$), COCO photographs ($n=10{,}000$, $0.9895$), and non-semantic conditioning ($n=960$, $0.8070$). No deployment threshold is selected.}
\label{fig:energy_roc}
\end{figure}

Each directional difference is zero at its corresponding image boundary. Higher scores indicate positive carriers. AUC is computed from ranks, with tied pairs counting as one half; the direction is not reversed for any negative set. Every comparison uses the same positive outputs. Figure~\ref{fig:energy_roc} shows the empirical ROC curves obtained by thresholding these saved scores. Lower discrimination on non-semantic conditioning indicates sensitivity to imposed non-semantic structure. This diagnostic does not measure correct content recognition or deployment precision.

\clearpage
\subsection{Qualitative Recovery}
\label{app:qualitative}

\begin{figure}[H]
\begin{center}
\includegraphics[width=\linewidth]{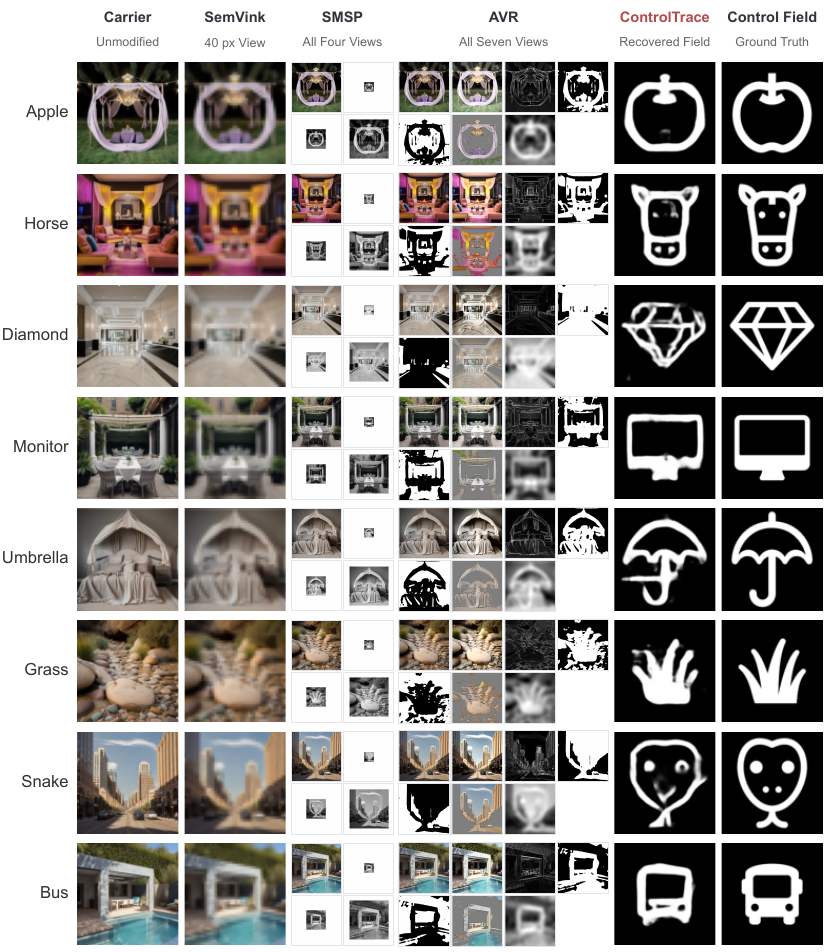}
\end{center}
\caption{All eight sampled targets, using the same columns as Figure~\ref{fig:teaser}, which shows the first three. Diamond has missing facets, and Snake is recovered as a different shape.}
\label{fig:negative}
\end{figure}

Figure~\ref{fig:negative} shows the complete sampled comparison underlying Figure~\ref{fig:teaser}, including its weaker recoveries. Figure~\ref{fig:matched} instead compares configuration-matched positive and negative generations; their purpose is to examine recovery when the conditioning branch is removed.

\end{document}